\documentclass[sigconf, nonacm]{acmart}
\usepackage[T1]{fontenc}
\usepackage{epstopdf}
\usepackage{graphicx}
\usepackage{float}
\usepackage{nomencl}
\usepackage{balance} 
\usepackage{color,soul}
\usepackage{multirow}
\usepackage{makecell}
\usepackage{longtable}
\usepackage{hhline}
\usepackage{pifont}
\usepackage{tabularx}
\usepackage[inline,shortlabels]{enumitem}
\usepackage{geometry}
\usepackage{xspace}
\usepackage[binary-units,per-mode=symbol,exponent-product=\cdot,list-units=single,list-final-separator={,}]{siunitx}
\DeclareSIUnit[number-unit-product = ]\pixel{p}
\DeclareSIUnit[number-unit-product = ]\dB{dB}
\usepackage{chemformula}
\usepackage{gensymb}
\usepackage{amsmath}
\usepackage{booktabs}
\usepackage{comment}
\usepackage{footnote}
\usepackage{tablefootnote}
\usepackage{caption}
\usepackage{subcaption} 
\usepackage{wrapfig}
\usepackage[linesnumbered,ruled,vlined]{algorithm2e}
\usepackage[noend]{algpseudocode}
\usepackage{xurl}
\usepackage{orcidlink}
\usepackage{hyperref}
\usepackage{hyperxmp}
\usepackage{footmisc}
\hypersetup{pdfauthor=author}
\newcommand{\etal}{\emph{et al.\xspace}}

\newcommand{\ie}{\emph{i.e.}, }
\newcommand{\eg}{\emph{e.g.}, }

\newcommand{\rf}[1]{\textcolor{black}{#1}}

\newcommand{\numbercircle}[1]{\textcircled{\scriptsize{#1}}}

\begin{document}
% \settopmatter{printfolios=false}
\title{ELLMPEG: An Edge-based Agentic LLM Video Processing Tool}

\author{Zoha Azimi}
\email{zoha.azimi@aau.at}
\affiliation{%
  \institution{Christian Doppler Laboratory ATHENA, Department of Information Technology (ITEC)\\ University of Klagenfurt}
  \city{Klagenfurt}
  \country{Austria}
}

\author{Reza Farahani}
\email{reza.farahani@aau.at}
\affiliation{%
  \institution{Department of Information Technology (ITEC)\\
  University of Klagenfurt}
  \city{Klagenfurt}
  \country{Austria}
}

\author{Radu Prodan}
\email{radu.prodan@uibk.ac.at}
\affiliation{%
  \institution{Department of Computer Science\\ University of Innsbruck}
  \city{Innsbruck}
  \country{Austria}
}

\author{Christian Timmerer}
\email{christian.timmerer@aau.at}
\affiliation{%
  \institution{Christian Doppler Laboratory ATHENA, Department of Information Technology (ITEC)\\ University of Klagenfurt}
  \city{Klagenfurt}
  \country{Austria}
}

\begin{abstract}
Large language models (LLMs), the foundation of generative AI systems like ChatGPT, are transforming many fields and applications, including multimedia, enabling more advanced content generation, analysis, and interaction. However, cloud-based LLM deployments face three key limitations: high computational and energy demands, privacy and reliability risks from remote processing, and recurring API costs. Recent advances in agentic AI, especially in structured reasoning and tool use, offer a better way to exploit open and locally deployed tools and LLMs. 
This paper presents \texttt{ELLMPEG}, an edge-enabled agentic LLM framework for the automated generation of video-processing commands. \texttt{ELLMPEG} integrates tool-aware \emph{Retrieval-Augmented Generation} (RAG) with iterative self-reflection to produce and locally verify executable \emph{FFmpeg} and \emph{VVenC} commands directly at the edge, eliminating reliance on external cloud APIs. To evaluate \texttt{ELLMPEG}, we collect a dedicated prompt dataset comprising \num{480} diverse queries covering different categories of \textit{FFmpeg} and the \textit{Versatile Video Codec} (VVC) encoder (VVenC) commands. We validate command generation accuracy and evaluate four open-source LLMs based on command validity, tokens generated per second, inference time, and energy efficiency. We also execute the generated commands to assess their runtime correctness and practical applicability. Experimental results show that \textit{Qwen2.5}, when augmented with the \texttt{ELLMPEG} framework, achieves an average command-generation accuracy of \qty{78}{\percent} with zero recurring API cost, outperforming all other open-source models across both the FFmpeg and VVenC datasets.

\end{abstract}

\setcopyright{none}
\acmConference{}{}{}
\acmBooktitle{}
\settopmatter{printacmref=false}

\maketitle
\section{Introduction}
\label{sec:intro}
In recent years, video traffic has dominated the Internet~\cite{sandvine}, requiring efficient processing tools to handle streaming and content creation~\cite{farahani2023sarena,farahani2022richter}. Among all multimedia processing tools, FFmpeg~\cite{ffmpeg} is widely used for video encoding~\cite{menon2023energy}, transcoding~\cite{menon2023transcoding}, and analysis, yet its complex command-line interface presents a challenge, driving developers and researchers to rely on different documentation or Artificial Intelligence (AI) assistants like ChatGPT for guidance. The rapid advancement of AI has driven extensive adoption of large language models (LLMs) across various fields, including multimedia~\cite{farahani2024towards}. Recently, both industry and academia have increasingly leveraged LLMs for various video processing tasks, such as content processing~\cite{long2024videodrafter,huang2024vtimellm,fu2024video, azimi2025towards}, video generation~\cite{wang2024lave,huang2024free}, and question-answering~\cite{lin2024streamingbench, zhang2023simple, pan2023retrieving}. For example, LLMPEG~\cite{llmpeg} was developed as an AI-powered solution that leverages OpenAI~\cite{openai} models to translate natural language queries into FFmpeg~\cite{ffmpeg} commands, providing users with an intuitive interface to simplify complex video processing tasks and unlock advanced functionalities.

\begin{figure*}
    \centering
    \includegraphics[width=.95\linewidth]{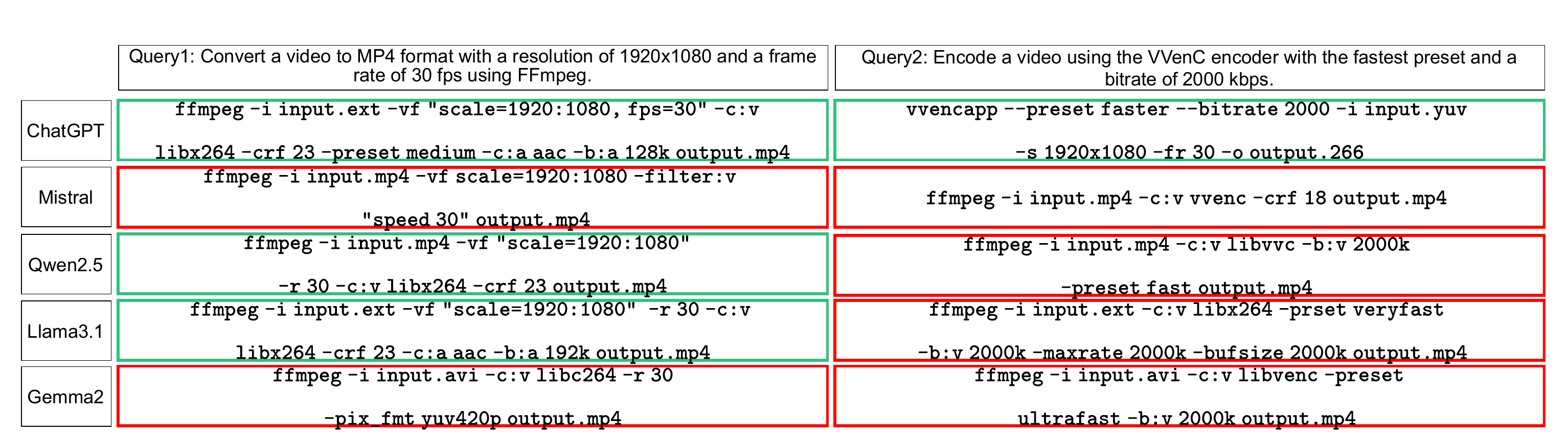}
    \caption{Comparison of responses to two queries: green borders indicate valid commands, red borders denote invalid ones.}
    \label{fig:fig1}
\end{figure*}

%%%
While cloud-based LLMs such as GPT-4o offer convenience and strong accuracy, they introduce several limitations: (1) \textit{network dependency}, which restricts offline operation in bandwidth-constrained environments; (2) \textit{recurring API costs} that scale with usage; and (3) \textit{limited customization}, especially for specialized tools whose documentation is not represented in cloud-model training data. These limitations motivate the use of edge-enabled, open-source, smaller (\numrange{2}{8} billion parameters) LLMs that execute on on-premise devices and operate with limited reliance on external cloud services.
However, smaller models face critical challenges in domain-specific command generation. As shown in Fig.~\ref{fig:fig1}, while models like Qwen2.5 (\num{7}B) and Llama3.1 (\num{8}B) perform reasonably well on common FFmpeg queries, they struggle with newer tools like VVenC, often hallucinating incorrect codec names (\textit{libvvc}, \textit{vvenc}) or non-existent command parameters. 
This accuracy–efficiency gap highlights the need for augmentation methods that enhance the performance of smaller, edge-deployable models while preserving the privacy and cost benefits of local execution.

Recent advancements in agentic LLMs introduce mechanisms for autonomous reasoning and iterative refinement~\cite{xi2025rise}. Approaches such as Reflexion~\cite{shinn2023reflexion}, Critic~\cite{gou2023critic}, and Self-Refine~\cite{madaan2023self} enable models to evaluate outputs through structured feedback loops, offering a principled way to enhance smaller edge-deployable models without increasing their parameter count. In the context of video processing, where precise command syntax and parameter selection are essential, these self-correcting mechanisms reduce hallucinations and substantially improve the reliability of generated commands. While most agentic frameworks rely on large foundational or cloud-based LLMs, recent work shows that smaller models can achieve competitive performance on specialized tasks, positioning them as a promising direction for practical agentic AI~\cite{belcak2025small}. 

This paper leverages agentic LLM reasoning for domain-specific multimedia processing command generation and introduces \texttt{ELLMPEG}, an edge-deployable agentic LLM system that produces reliable FFmpeg and VVenC video-processing commands. By integrating Retrieval-Augmented Generation (RAG) with self-reflection mechanisms, \texttt{ELLMPEG} narrows the accuracy gap between edge-deployable models and large cloud LLMs while preserving the cost advantages of local execution. \texttt{ELLMPEG} implements a multi-stage workflow comprising (1) query interpretation and processing tool (\eg FFmpeg, VVenC, or both) selection, (2) retrieval of tool-specific documentation, (3) command generation, (4) self-critique to identify errors or hallucinations, and (5) iterative refinement until a correct FFmpeg or VVenC command is produced. The main contributions of this work are as follows:
\begin{enumerate}[topsep=0pt]
    \item \textit{Design} an edge-deployable agentic LLM architecture with a minimal-overhead self-reflection loop tailored for resource-constrained 2–8B-parameter models.
    \item \textit{Develop} a tool-aware RAG design with dual vector stores that index FFmpeg and VVenC documentation for precise multimedia command generation.
    \item \textit{Collect} a comprehensive dataset of \num{480} FFmpeg and VVenC queries for benchmarking command generation systems\footnote{https://github.com/zoha-az/ELLMPEG}.
    \item \textit{Provide} an extensive evaluation of open-source models on correctness, performance, and energy efficiency.
\end{enumerate}

This paper has six sections. Section~\ref{sec:sota} reviews state-of-the-art and highlights \texttt{ELLMPEG} novelties. Section~\ref{sec:arch} designs the \texttt{ELLMPEG} system and describes its architecture and algorithms. Section~\ref{sec:eval_set} details the dataset and evaluation setup, followed by the evaluation results in Section~\ref{sec:res}. Finally, Section~\ref{sec:conclusion} concludes the article. 
%%%%%

\section{Related Work}
\label{sec:sota}
\sloppy
\begin{table*}[!t]
\centering
\caption{Related work summary (MM: Multimedia).}
\label{tab:sota}
\scriptsize
\begin{tabular}{|c|c|c|c|c|c|}
\toprule
\multirow{2}{*}{\textbf{Category}} & \multirow{2}{*}{\textbf{Paper}} & \multirow{2}{*}{\textbf{Year}} & \multirow{2}{*}{\textbf{Base model}}  & \multirow{2}{*}{\textbf{Use case}}   & \textbf{LLM enhancement}     \\
&  &  &   &   & \textbf{technique}   \\
\toprule
%%%%
\multirow{7}{*}{MM processing} &\multirow{2}{*}{LAVE~\cite{wang2024lave}}   & \multirow{2}{*}{2024} & LLaVA-v1.0~\cite{liu2023visual}  & Video editing& \multirow{2}{*}{\ding{53}}       \\ 
& &  & + GPT-4   &   (video trimming, sequencing) &  \\ \cline{2-6}
 %%%%
&LLMPEG~\cite{llmpeg} & 2024 & OpenAI models (GPT-4)   & FFmpeg command generation   & \ding{53}   \\ \cline{2-6}
%%%
 & RAVA~\cite{cao2024reframe} & 2024 & OpenAI models (GPT-4) &  Video reframing & \ding{53} \\ \cline{2-6}
%%%
& M2-Agent\cite{tran2025towards} & 2025 & Llama3.3-70B & Video object segmentation & Agentic (interaction with tools)
 \\ \cline{2-6}
& Prompt-Driven Agentic & \multirow{2}{*}{2025} & \multirow{2}{*}{Gemini 2.0 Flash~\cite{gemini}} & \multirow{2}{*}{Video editing} & \multirow{2}{*}{Agentic (orchestration of tools)} \\ 
& Video Editing System\cite{ding2025prompt} &  &  &  \\ \hline

%%%% 
\multirow{3}{*}{MM generation} &VideoStudio~\cite{long2024videodrafter} & 2024 & ChatGLM3-6B~\cite{du2021glm}, GPT-4 & Multi-scene video generation  & \multirow{1}{*}{\ding{53}}       \\ \cline{2-6}
%%%%
&Free-Bloom~\cite{huang2024free}   & 2024 & ChatGPT  & Semantically coherent video generation   &  \ding{53}  \\ \cline{2-6}
%%%%
&VideoDirectorGPT~\cite{lin2023videodirectorgpt}  & 2023 & GPT-4  & Multi-scene video generation   &  \ding{53}  \\ \hline

%%%%
\multirow{8}{*}{MM understanding} &\multirow{2}{*}{Video-RAG~\cite{luo2024video}} & \multirow{2}{*}{2024} & Open source models (\eg & \multirow{2}{*}{Video question answering  } &  \multirow{2}{*}{RAG}       \\ 
&&  & Video-LLaVA\cite{lin2023video}, Qwen2-VL\cite{wang2024qwen2})&    &   \\ \cline{2-6}
%%%%
&ViTA~\cite{arefeen2024vita} & 2024 &   gpt-3.5-turbo~\cite{openai}  & Video question answering   &  RAG   \\ \cline{2-6}
%%%%
&iRAG~\cite{arefeen2024irag}      & 2024 &  gpt-3.5-turbo~\cite{openai}      &   Video question answering   &   RAG      \\ \cline{2-6}
%%%%
&\multirow{2}{*}{VTimeLLM~\cite{huang2024vtimellm}}        & \multirow{2}{*}{2024} & \multirow{2}{*}{Vicuna~\cite{vicuna2023}}  &  Time-bound      &   \multirow{2}{*}{Fine-tuning} \\ 
&         &  &    &     video question answering    &         \\ \cline{2-6}
&\multirow{1}{*}{AVA~\cite{yanava}}        & \multirow{1}{*}{2025} & \multirow{1}{*}{Qwen2-VL\cite{wang2024qwen2}}  &  Long video question answering     &   \multirow{1}{*}{Agentic LLM with RAG} \\ \cline{2-6}
& Deep Video Discovery~\cite{zhang2025deep}& 2025 &  OpenAI models (GPT-4) & Long video question answering & Agentic (interaction with tools) \\ \bottomrule
%%%%
\multirow{2}{*}{MM processing} &\multirow{2}{*}{\texttt{ELLMPEG}}        & \multirow{2}{*}{2026} & Open source models  &    FFmpeg and VVenC    &   \multirow{2}{*}{ Agentic LLM + RAG + self-refinement} \\ 
&         &  & (\eg Qwen2.5~\cite{qwen}, Gemma2~\cite{team2024gemma})    &     command generation    &         \\ \bottomrule
\end{tabular}
\end{table*}
Integrating natural language processing with general-purpose programming tasks has emerged as one of the most impactful applications of LLMs, extensively explored in recent research~\cite{aggarwal2025programming, ahmed2024codeqa, alario2024tailoring, nam2024using}, and supported by a rich ecosystem of datasets, benchmarks, and specialized fine-tuned models. However, such progress has not yet extended to multimedia domains, where powerful tools like FFmpeg remain largely inaccessible to non-expert users. Recent advancements demonstrate the potential of integrating LLMs into multimedia domains to simplify complex tasks, with state-of-the-art research primarily focusing on three key areas: \textit{(a)} processing, \textit{(b)} generation, and \textit{(c)} understanding.
%%%
\subsection{Multimedia processing}
Wang~\etal~\cite{wang2024lave} developed a video editing interface, allowing users to process and perform editing tasks, such as sequencing and trimming, and generate video clips using a gallery of video sequences. They employed LLaVA (Large Language and Vision Assistant)-v1.0~\cite{liu2023visual} to generate textual descriptions of the video sequences, enabling the retrieval of clips relevant to the user's specified topic. They also used GPT-4 to process the user's natural language commands, translating them into FFmpeg editing instructions to modify the selected clips accordingly.
LLMPEG~\cite{llmpeg} provides a command-line tool that translates natural language instructions into precise FFmpeg commands, automating multimedia processing. It leverages OpenAI models such as GPT-4o to generate and execute platform-aware commands, adapting to user queries and FFmpeg configurations while enabling interactive refinement for greater accessibility.
Cao~\etal~\cite{cao2024reframe} presented RAVA, a system that automates video reframing by leveraging LLMs such as GPT-4.
RAVA interprets user instructions expressed in natural language, identifies key objects or scenes within the video, and automatically adjusts the framing and aspect ratio to emphasize those elements.

Tran~\etal~\cite{tran2025towards} leveraged Llama-3.3-70B~\cite{llama3.3} to develop M2-Agent, an agentic system that plans and guides object segmentation in video sequences using multimodal cues such as text and audio. The system dynamically constructs tailored workflows via step-by-step reasoning, calling specialized tools such as Grounding DETR with improved denoising anchor
boxes (DINO)~\cite{liu2024grounding} for object detection, Segment Anything Model (SAM)-2~\cite{ravi2024sam} for video segmentation, and Bidirectional Encoder representation from Audio Transformers (BEATs)~\cite{chen2022beats} for audio classification to handle low-level multimodal processing and accurately locate and segment target objects in a video sequence without needing task-specific training. Ding~\etal~\cite{ding2025prompt} present a prompt-driven agentic video-editing framework that autonomously interprets and restructures long-form narrative media via modular agents orchestrated through natural-language prompts. The system leverages Gemini 2.0 Flash to process media, produce storyboards, refine scene descriptions, and guide tasks, including clip retrieval and narrative planning.
%%%
\subsection{Multimedia generation}
Several works~\cite{long2024videodrafter, huang2024free} leveraged LLMs to generate video sequences from textual descriptions. Lin~\etal~\cite{lin2023videodirectorgpt} proposed a framework that leverages GPT-4 to transform a single text prompt into a detailed video plan, detailing scene descriptions, spatial layouts, and background to 
guide a video generator, ensuring spatial control and temporal consistency across multiple scenes.  
Long~\etal~\cite{long2024videodrafter} used ChatGLM-3-6B~\cite{du2021glm} and GPT-4 to convert input prompts into detailed scene descriptions, background and foreground elements, and camera movements. These descriptions generate reference images via a text-to-image model, which a diffusion model leverages to produce a multi-scene video. Similarly, Huang~\etal~\cite{huang2024free} also used ChatGPT to break down an input prompt into a sequence of detailed descriptions capturing content evolution over time, guiding a diffusion model in generating coherent frames.

\subsection{Multimedia understanding}
Recent work has primarily applied RAG to video-understanding and video question-answering pipelines~\cite{arefeen2024vita, arefeen2024irag}. 
Luo~\etal~\cite{luo2024video} proposed Video-RAG to improve the understanding of long video sequences by augmenting large video-language models with auxiliary data, including audio and object detection outputs, incorporated alongside the video frames and user queries.
Huang~\etal~\cite{huang2024vtimellm} fine-tuned the Vicuna~\cite{vicuna2023} model with annotated datasets to enhance its ability to describe specific video segments and perform temporal reasoning, enabling precise identification of events with accurate start and end boundaries. Yan~\etal~\cite{yanava} introduced AVA, a vision-language model-powered system for open-ended, video understanding that can process ultra-long and continuous video streams. It generates event knowledge graphs to efficiently index long videos and employs an agentic retrieval-generation mechanism to answer complex queries by dynamically exploring and synthesizing relevant video event information.
Zhang~\etal~\cite{zhang2025deep} introduced an agentic system that uses GPT-4.1 to autonomously analyze long videos by iteratively searching for relevant information through specialized tools at multiple granularities, i.e., in a global, segment, or frame level. The LLM functions as a cognitive driver, reasoning, planning, and deciding which tools to use and when, based on accumulated observations, to effectively decompose complex queries.
\subsection{State-of-the-art limitations}
Table~\ref{tab:sota} summarizes state-of-the-art works, detailing the LLM used, the target use case, and whether the model was applied directly for inference or enhanced via fine-tuning, RAG or agentic design. Most LLM-based video-processing tools depend either on general-purpose cloud models like ChatGPT or Gemini, which restrict customization and accessibility, or on large-scale models that are infeasible to deploy on edge hardware. To our knowledge, \texttt{ELLMPEG} is the first agentic LLM-based video processing tool to leverage open-source models and a customized RAG framework while efficiently responding to queries.

\section{\texttt{ELLMPEG} Design}
\label{sec:arch}
\begin{figure*}[t!]
    \centering    
\includegraphics[width=1\linewidth]{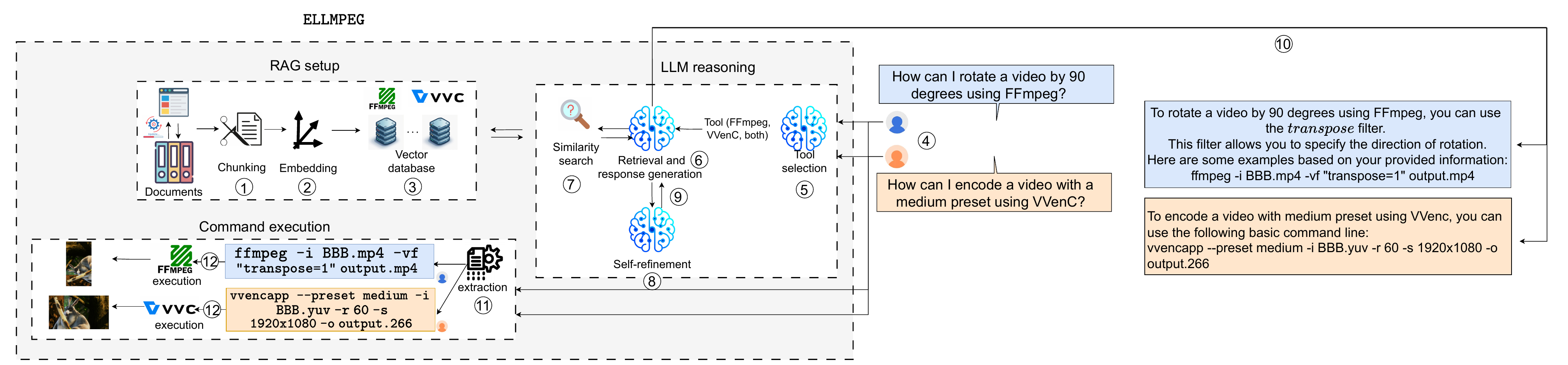}
    \caption{\texttt{ELLMPEG} architecture.}
    \label{fig:arch}
\end{figure*}

The \texttt{ELLMPEG} architecture shown in Fig.~\ref{fig:arch} comprises three phases: \textit{(a)} RAG setup, \textit{(b)} LLM reasoning, and \textit{(c)} command execution. Table~\ref{tab:notation} summarizes the notations used in the algorithm.
%%%%%%%%%
%%%%%%%%
\subsection{RAG setup}
This phase processes documents relevant to user queries to improve the LLM's query-answering capabilities. \texttt{ELLMPEG} executes this phase once and updates it incrementally upon detecting changes in the documentation, following the steps \numbercircle{1}--\numbercircle{3} outlined in Fig.~\ref{fig:arch}.
%%%
\begin{table}[t!]
\caption{Important notations.}
\label{tab:notation}
\scriptsize
\begin{tabular}{|c|c|c|c|}
\hline
\textbf{Notation}      & \textbf{Description}     & \textbf{Notation}   & \textbf{Description}    \\ \hline
$s$ & Target chunk size &  $l$ & Max. document size\\ \hline 
$\mathcal{D}$ & List of  delimiters & $\mathcal{C}$ & Chunks description\\ \hline
$k$&   Number of chunks to return   &   $\mathcal{DC}$ &  Documents     \\  \hline
$\mathcal{CH}$ &  Chunks  &     $\mathcal{QV}$ &  Query vector     \\  \hline
$I_{\max}$ & Max. iteration    & $\mathcal{SS}$ &  Similarity score     \\  \hline
$\mathcal{T}$  & Top $k$ chunks     & $\mathcal{Q}$ & User query \\ \hline
$\mathcal{M}$ & Language model & $E$ & Embedding model \\ \hline
$\mathcal{VS}_{f}$ & FFmpeg vector store & $\mathcal{VS}_{v}$ & VVenC vector store \\ \hline
$\mathcal{CV}_{f}$ & FFmpeg chunk vector & $\mathcal{CV}_{v}$ & VVenC chunk vector \\ \hline
$\mathcal{A}$ & Generated response & $\mathcal{F}$ & Self-reflection feedback \\ \hline
\end{tabular}
\end{table}
%%%
\subsubsection{Chunking} 
Chunking processes raw documents ($\mathcal{DC}$), segmenting them into smaller, manageable units called \emph{chunks} ($\mathcal{CH}$) to facilitate efficient indexing and retrieval. 
As detailed in Alg.~\ref{alg:ellmpeg_full}, it takes two additional configurable inputs alongside $\mathcal{DC}$: (1) chunk size ($s$) defines the maximum length of each chunk and (2) a set of delimiters ($\mathcal{D}$), an ordered list of characters, such as spaces, punctuation marks, or newlines, that define valid chunking points within each document. The output is a list of chunk descriptions ($\mathcal{C}$), each containing the textual content of a chunk $ch\in\mathcal{CH}$ and its associated metadata $m$, which includes the document $dc\in\mathcal{DC}$ from which the chunk was extracted. It also outputs a tool tag indicating whether the chunk pertains to \texttt{FFmpeg} or \texttt{VVenC}, allowing the system to route retrievals to the correct tool-specific vector store.

First, it initializes $\mathcal{CH}$ and $\mathcal{C}$, along with an auxiliary parameter $d$, which indicates the position of the first delimiter in $\mathcal{D}$ (line 1). Next, it iterates through all documents in $\mathcal{DC}$ and invokes the \texttt{Split} function to generate chunks $\mathcal{CH}$ (lines 2-3). For each chunk $ch$, the algorithm evaluates its size, updates the delimiter pointer $d$, and recursively calls the \texttt{Split} on $ch$ to achieve a better granularity (lines 4–7). It then invokes the \texttt{Extraction} function for each chunk $ch$ from document $dc$ to generate its metadata $m$ (line 8). It subsequently updates $\mathcal{C}$ with that chunk content and metadata (line 9) and returns $C$ for all documents. 
\subsubsection{Embedding}
\texttt{ELLMPEG} maintains two separate embedding vector stores, one for FFmpeg ($\mathcal{CV}_f$) and one for VVenC ($\mathcal{CV}_v$) (line 10), to leverage the tool-aware metadata generated during chunking. The embedding function \texttt{Embed} analyzes chunks in $\mathcal{C}$, capturing semantic and syntactic relationships within the text and converting them into numerical vectors (lines 11-12). This transformation enables computationally efficient search and retrieval by representing text as dense numerical embeddings, rather than processing raw text. The embedding vectors for each tool are then stored in a different vector (lines 13-16). 
\subsubsection{Vector database}
The vector database computes embeddings for efficient retrieval using flat indexing, storing them in a structured array that enables direct similarity comparisons and precise search results. The embedding vectors for each tool are stored in separate FAISS vector stores, $\mathcal{VS}_{f}$ for FFmpeg and $\mathcal{VS}_{v}$ for VVenC (line 17). This separation reduces the search space by directing queries only to the relevant store and eliminates cross-tool retrieval noise, preventing irrelevant documentation from contaminating the retrieved context.
\subsection{LLM reasoning}
This phase processes the user query and generates a response, following the steps \numbercircle{4}--\numbercircle{10} in Fig.~\ref{fig:arch}.
When the user issues a query, \eg``\textit{How can I rotate a video by 90 degrees using FFmpeg?}'' (\numbercircle{4}), the system proceeds through the following steps.
%%%
\subsubsection{Tool selection}
The LLM $\mathcal{M}$ determines which video processing tool (i.e., FFmpeg, VVenC, or both) is required to answer the query. The model $\mathcal{M}$ is prompted to select the appropriate tool based on the query content, producing a single-word label $t$ (FFmpeg, VVenC, or both) (line 18). This label determines which vector store(s) will be queried in the subsequent step.
%%%%
\subsubsection{Response generation with retrieval}
This step converts the user query into a vector using the same embedding strategy \numbercircle{2}, ensuring that the query and the document chunks reside in the same vector space and enabling effective similarity comparisons (line 19). 
To retrieve relevant context, the system applies a similarity search over the selected vector store(s) ($\mathcal{VS}_{f}, \mathcal{VS}_{v}$), returning the top-$k$ most relevant chunks $\mathcal{T}$ for the user query $\mathcal{Q}$.
It starts this process by initializing $\mathcal{SS}$ and $\mathcal{T}$ to store similarity scores and top chunks (line 20).
It then iterates over all chunk vectors in $\mathcal{VS}_{f}, \mathcal{VS}_{v}$ (line 21) and over all elements of each chunk vector (line 23) to compute the \textit{Euclidean} distance (lines 24 - 25). Finally, it invokes the heap-based \texttt{Sort} function to rank the chunks by their similarity scores $\mathcal{SS}$ and select the top-$k$ most relevant ones in the set $\mathcal{T}$ (line 26).
Using the retrieved chunks $\mathcal{T}$ and the original query $\mathcal{Q}$, the LLM $\mathcal{M}$ generates a detailed response, which typically includes a complete command (e.g., ``\texttt{ffmpeg -i input.mp4 -vf 'transpose=1' output.mp4}'') along with explanations of what the command does and why as the answer $\mathcal{A}$ (line 27). 
 
\subsubsection{Self-reflection and iterative revision}
To enhance accuracy and completeness, the generated response undergoes an iterative self-reflection loop. As long as the maximum iteration threshold $I_{max}$ has not been reached (line 29), $\mathcal{M}$ evaluates its own output and produces feedback $\mathcal{F}$ regarding correctness, completeness, syntax, and clarity (line 30), conditioned on the input query $\mathcal{Q}$ and the generated response $\mathcal{A}$. If the feedback indicates that revisions are needed, the LLM revises the response accordingly. This loop continues for a maximum of $I_{\max}$ iterations or until no further revisions are required (lines 32-33).
%%%
\subsection{Command execution}
The final phase of the \texttt{ELLMPEG} pipeline facilitates the transition from LLM-generated text (step \numbercircle{10}) to executable commands for FFmpeg and VVenC, bridging natural language interaction with real-world, verifiable system actions. Once the LLM generates a full-text response, a lightweight command extraction module scans the response for executable content (\numbercircle{11}). 
% \za{In this work, we did not force structured outputs on $\mathcal{M}$s, since models such as Gemma with 2B parameters struggle with such constraints. Instead, }
\texttt{ELLMPEG} applies pattern-matching rules using regular expressions to detect and isolate shell-compatible commands (e.g., \texttt{ffmpeg -i ... -vf "transpose=1"}), discarding verbose explanations or non-functional text. The extracted commands $\mathcal{CMD}$ are then passed to the execution script (step \numbercircle{12}), which dispatches them to appropriate backends such as FFmpeg or VVenC for actual processing.
The result of execution is a transformed media output (e.g., \texttt{``rotated or encoded video''}), which can be optionally visualized or benchmarked.
%%%
\subsection{Time complexity}
The overall complexity of \texttt{ELLMPEG} is dominated by chunking and similarity search. The chunking runs in $O(|\mathcal{DC}| \cdot l)$, where $l$ is the maximum document length, due to recursive splitting. The similarity search dominates retrieval with $O(N\cdot d)$ per query, where $N$ is the number of chunks and $d$ is the embedding dimension. The self-reflection step incurs up to $I_{\max}$ additional model inferences but does not change the complexity.

\section{Evaluation Setup}
\label{sec:eval_set}
%%%%
\begin{figure*}
    \centering
\includegraphics[width=.65\linewidth]{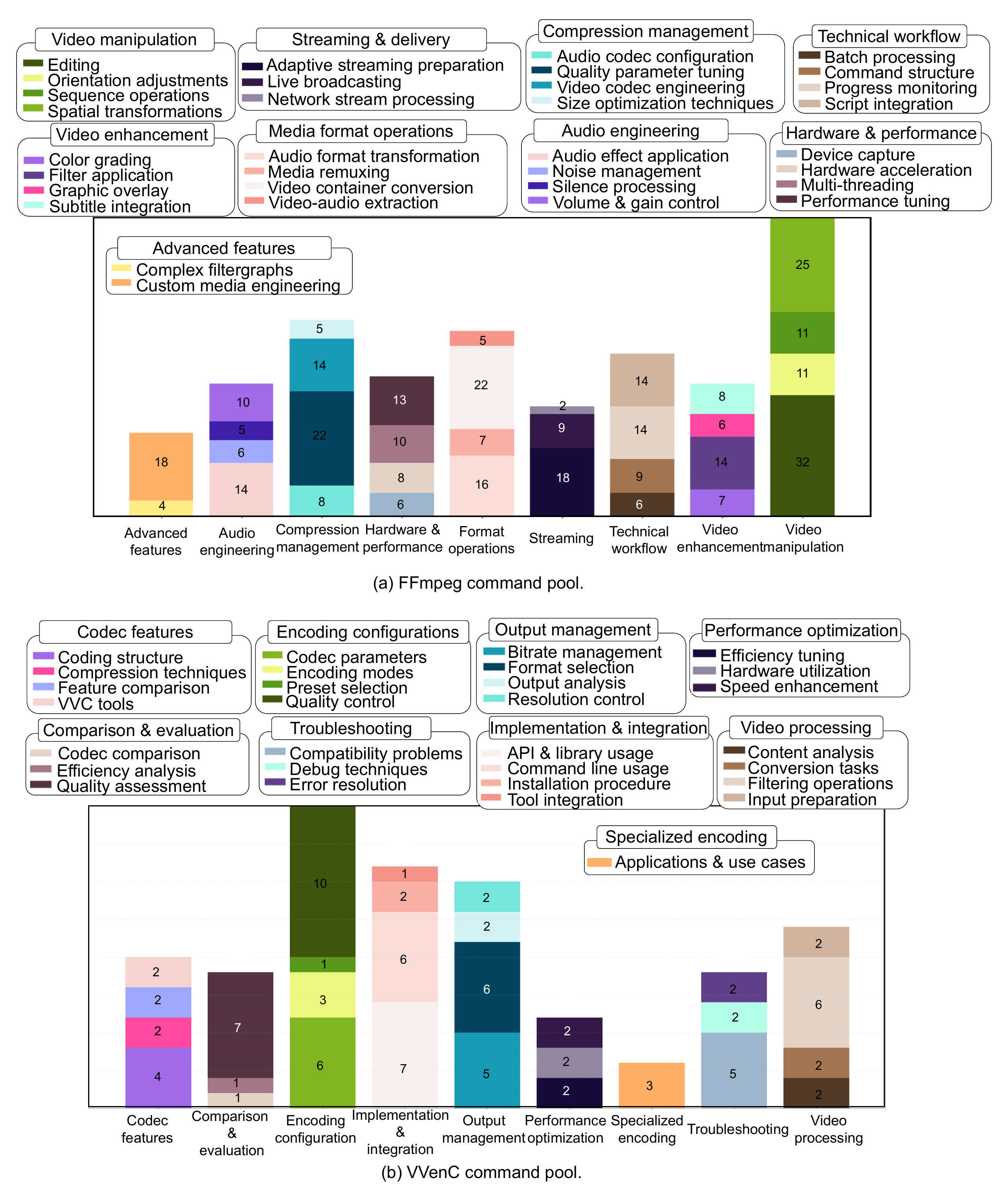}
    \caption{Query distribution across various categories in the (a) FFmpeg and (b) VVenC command pools.}
    \label{fig:pool}
\end{figure*}
%%%
Our edge evaluations were performed on an Intel i7-8700 CPU system, while server experiments used a 128-core Intel Xeon Gold machine with two NVIDIA Quadro GV100 GPUs. Since \texttt{ELLMPEG} does not rely on any hardware-specific optimizations or GPU-dependent components, it can be executed on other CPU-based edge devices with similar capabilities.
The following subsections detail the test set command pools, documents used in the RAG setup phase, configurations, and evaluation metrics.
%%%
\subsection{Test set command pool}
We constructed two distinct command pools of \num{480} diverse multimedia processing queries to evaluate \texttt{ELLMPEG}, filling the gap left due to the lack of an existing dataset. We adopted a hybrid approach by integrating commands from multiple reliable sources, including \num{200} LLM-generated queries using \texttt{GPT-4o} for diverse, intent-driven scenarios, and \num{280} real-world examples collected from GitHub repositories~\cite{gitffmpeg, gitvvc1, gitvvc2}, code snippets, and usage patterns from the community forums~\cite{wikiffmpeg, wikivvc}.
The dataset comprises \num{380} FFmpeg and \num{100} VVenC queries, each systematically categorized to capture practical use cases and operational diversity, as shown in Fig.~\ref{fig:pool}. 
The query distribution ($\num{380}$ for FFmpeg and $\num{100}$ for VVenC) reflects the functional diversity of each tool. FFmpeg is a comprehensive multimedia framework integrating multiple codec libraries (e.g., AVC~\cite{wiegand2003overview}, HEVC~\cite{sullivan2012overview}), requiring a larger corpus to capture its breadth. In contrast, VVenC is a specialized standalone encoder with a narrower configuration space, for which $\num{100}$ queries sufficiently cover typical encoding scenarios. \texttt{ELLMPEG} further benefits from a tool-agnostic design. While our evaluation focuses on FFmpeg and VVenC, the framework readily extends to other formats or hardware-specific encoders. Thanks to its RAG-based architecture, adaptation only requires updating the tool documentation in the knowledge base, without retraining the underlying LLM.
To ensure reproducibility and support further research, both FFmpeg and VVenC command pools are publicly released\footnote{https://github.com/zoha-az/ELLMPEG}.  
%%%
\begin{algorithm}[!t]
\caption{\texttt{ELLMPEG}: End-to-end agentic retrieval-augmented multimedia processing system.}
\footnotesize
\label{alg:ellmpeg_full}
\KwIn{$\mathcal{DC}$, $s$, $\mathcal{D}$, $E$, $\mathcal{M}$, $\mathcal{Q}$, $I_{\max}$, $k$} 
\KwOut{$\mathcal{A}$}
% =======================
% STEP 1: CHUNKING
% =======================
\tcp{Step 1: Document chunking and metadata extraction} 
$\mathcal{CH} \gets \varnothing$,
$\mathcal{C} \gets \varnothing$, $d\gets 0$ \\
\For{$ dc \in \mathcal{DC}$} {
    $\mathcal{CH} \gets \texttt{Split}(dc, d)$        \\
    \For{$ ch \in \mathcal{CH}$} {
    \While{$|ch| > s $} {
        $ d \leftarrow d+1 $\\
        $ch \gets \texttt{Split}(ch, d)$ 
    }         
    $m \gets \texttt{Extraction}(ch, dc)$  \\ 
    $\mathcal{C}\gets (ch, m)$\\  
}}
% =======================
% STEP 2: EMBEDDINGS & VECTOR STORE CREATION
% =======================
\tcp{Step 2: Embedding and vector store construction} 
$\mathcal{CV}_{f} \gets \varnothing$, $\mathcal{CV}_{v} \gets \varnothing$ \\[2pt]

\For{$(ch, m) \in \mathcal{C}$}{
    $v \gets \texttt{Embed}(E, ch)$  \\ 
    \uIf{$m.tool == \texttt{FFmpeg}$}{
        $\mathcal{CV}_{f} \gets \mathcal{CV}_{f} \cup \{v\}$ 
    }\ElseIf{$m.tool == \texttt{VVenC}$}{
        $\mathcal{CV}_{v} \gets \mathcal{CV}_{v} \cup \{v\}$ 
    }
}
$\mathcal{VS}_{f} \gets \texttt{BuildFAISS}(\mathcal{CV}_{f})$ ,
$\mathcal{VS}_{v} \gets \texttt{BuildFAISS}(\mathcal{CV}_{v})$ 

% =======================
% STEP 3: TOOL SELECTION
% =======================
\tcp{Step 3: Tool Selection via LLM} 
$ $t$ \gets \mathcal{M}(\mathcal{Q})$ 

% =======================
% STEP 4: SIMILARITY SEARCH
% =======================
\tcp{Step 4: Response generation with retrieval} 
$\mathcal{QV} \gets \texttt{Embed}(E, \mathcal{Q})$ \\
$\mathcal{SS} \gets \varnothing , \mathcal{T} \gets \varnothing$\\
\For{$cv \in \mathcal{VS}_{f}|\mathcal{VS}_{v}$}{
    $dist \gets 0, i \gets 0$ \\
    \For{$i$ \KwTo $|cv|$}{
        $dist \gets (\mathcal{QV}[i] - cv[i])^2$
    }
    $\mathcal{SS} \gets \sqrt{dist}$ \\    
}
$\mathcal{T} \gets \texttt{Sort}(\mathcal{SS}, k)$ \\
$\mathcal{A} \gets \mathcal{M}(\mathcal{Q}, \mathcal{T}, t)$ \\

% =======================
% STEP 6: SELF-REFLECTION LOOP
% =======================
\tcp{Step 6: Self-reflection and revision} 
$i \gets 1$, $\mathcal{F} \gets \varnothing$ \\

\While{$i < I_{\max}$}{
    $\mathcal{F}_i \gets \mathcal{M}(\mathcal{Q}, \mathcal{A}, \mathcal{T})$ %\tcp*{Feedback for iteration i}
    \If{$\texttt{NeedsRevision}(\mathcal{F}_i) = \texttt{False}$}{
        \textbf{break}
    }
    $\mathcal{A} \gets \mathcal{M}(\mathcal{Q}, \mathcal{T}, \mathcal{A}, \mathcal{F}_i)$ \\  
    $i \gets i + 1$
}
\KwRet $\mathcal{A}$
\end{algorithm}
%%%

The FFmpeg queries cover the following categories:
\begin{enumerate*}
    \item\emph{Video manipulation}  includes all operations that structurally or visually alter video content, such as trimming, resizing, rotating, and concatenating.
    \item\emph{Compression management} involves configuring encoding parameters and compression techniques, including bitrate and constant rate factor (CRF) tuning, codec selection, and audio compression settings. 
    \item\emph{Media format} operations cover container and format transformations, such as changing video containers, and separating streams from multimedia files.
    \item\emph{Technical workflow} focuses on command structuring and execution pipelines, including status tracking, script integration, and batch processing for handling multiple files efficiently.
    \item\emph{Hardware and performance} focuses on resource utilization and optimization, including multi-threading, GPU acceleration, and hardware-based tasks such as screen or camera recording.
    \item\emph{Video enhancement} involves applying visual effects and modifications, such as adding text elements, watermarks, logos, and color adjustments. 
    \item\emph{Audio engineering} addresses audio processing tasks, including applying effects, managing silent segments, and reducing artifacts or noise.
    \item\emph{Streaming and delivery} focuses on content distribution techniques, including support for HLS and DASH formats, real-time streaming, and network-based media protocols. 
    \item\emph{Advanced features} encompass specialized and complex operations, such as custom media processing workflows and intricate filter configurations.
\end{enumerate*}

The VVenC queries cover the following categories:
\begin{enumerate*}
    \item\emph{Encoding configuration} includes settings and parameters that control the encoding process, e.g., quantization settings (e.g., QP values), rate control options, and preset selection.
    \item\emph{Output management}  focuses on configuring the characteristics of the encoded output, including target bitrate, resolution, and aspect ratio settings.
    \item\emph{Implementation and integration} covers the practical use of VVenC across different environments, including encoder setup and configuration, workflow integration, and command-line interface (CLI) syntax.
    \item\emph{Video processing} covers tasks related to handling video content, including preprocessing filters, format conversion, and analysis of video characteristics.
    \item\emph{Comparison and evaluation} involves benchmarking and quality assessment tasks, such as measuring output quality and evaluating compression efficiency.
    \item\emph{Troubleshooting} focuses on debugging and error resolution, including handling unsupported features, diagnosing encoding failures, and resolving command execution issues.
    \item\emph{Codec features} focuses on VVC-specific capabilities, including frame types, prediction structures, and Group of Pictures (GOP) configurations.
    \item\emph{Performance optimization} focuses on improving encoding speed and resource efficiency, including techniques to accelerate processing and configure CPU, GPU, and memory usage effectively.
    \item\emph{Specialized encoding} covers targeted use cases and application scenarios, such as live streaming, real-time encoding, and platform-specific requirements.
\end{enumerate*}
%%%
\begin{table*}[!t]
\caption{Pre-trained open-source LLMs used by \texttt{ELLMPEG}.}
\centering
\label{tab:llm}
\setlength{\tabcolsep}{1pt}
\begin{tabular}{|c|c|c|c|c|}
\hline
  \textbf{Model} &
  \textbf{Parameters (billions)} &
  \textbf{Specialization} &  
  \textbf{Context length (tokens)} &  \textbf{Data freshness} \\ \hline
  \textbf{Qwen2.5}~\cite{qwen} & 7 & Code generation, mathematics &   \num{128000} & Sep. 2023 \\ \hline   
  \textbf{Llama3.1}~\cite{touvron2023llama} &  8 & General-purpose  & \num{128000} &
  Dec. 2023  \\ \hline   
  \textbf{Gemma2}~\cite{team2024gemma} & 2 &   Code generation, symbolic reasoning &  \num{8000} &   Sep. 2023 \\ \hline    
  \textbf{Mistral}~\cite{jiang2023mistral} & 7 & General-purpose    &   \num{32000}   & Mid-2023  \\ \hline     
\end{tabular}
\end{table*}
%%%
\subsection{RAG documents} 
We extracted the documents for the RAG setup phase from the official FFmpeg and VVenC  documentation\footnote{\url{https://ffmpeg.org/documentation.html}}\textsuperscript{,}\footnote{\url{https://github.com/fraunhoferhhi/vvenc/wiki}}. We compiled nine PDFs from FFmpeg documentation~\cite{ffmpegDoc}, covering FFmpeg, ffprobe, and ffplay tools and components. These include command-line tools, codecs, and utilities, with details from the components documentation and frequently asked questions from the general documentation, ensuring comprehensive coverage of FFmpeg's core functionality and commonly encountered use cases. 
We created a consolidated PDF for VVenC from its wiki page~\cite{vvcDoc}, incorporating instructions on obtaining, building, and using VVenC, as well as additional topics like multi-threading and rate control.
%%%%%
\subsection{\texttt{ELLMPEG} configuration}
This section details \texttt{ELLMPEG} configuration, including the chunking parameters, embedding model, vector database library, similarity search method, and parameters such as the number of documents to return and retrieval parameters. 
%%%
\subsubsection{Chunking} 
We employed the \textit{RecursiveCharacterTextSplitter}~\cite{TextSplitter} method for the splitting process, taking chunking parameters $s$ and $\mathcal{D}$ into account. We set $s$ based on the context window of downstream LLMs, defined by their maximum token processing capacity.
We selected a context window of \num{4000} tokens for LLMs detailed in Table~\ref{tab:llm}, ensuring compatibility with their processing capabilities.
Since each token corresponds to approximately four English characters~\cite{tokentochar}, 
this translates to a processing limit of around \num{16000} characters. Consequently, each model handles up to \num{16000} characters at a time, encompassing both the query and the most relevant chunks.
We set $k=5$ to retrieve and process the five most relevant chunks of text (see~\ref{sim}). Given a total capacity of \num{16000} characters and the need to accommodate the query, we allocated approximately \num{3000} characters per chunk.

We applied an overlap of \num{500} characters between consecutive chunks to preserve continuity and avoid losing context. We set the delimiter list $\mathcal{D}$ based on a hierarchical text structure, prioritizing paragraphs (\ie double line breaks) first, followed by single line breaks (\ie new lines), then sentences identified by punctuation such as periods, exclamation marks, or question marks, and finally individual words. We finally associated each chunk with metadata, including the file name, which identifies its source for traceability, and the chunk index, which specifies its order within the file.
%%%
\subsubsection{Embedding} 
\rf{We used the compact version of bge-small-en-v1.5~\cite{bge_embedding}, a transformer-based embedding model by the Beijing Academy of Artificial Intelligence (BAAI), featuring \num{33.4} million parameters and optimized for efficient semantic text similarity and retrieval tasks on both CPU and GPU devices.}
%%%
\subsubsection{Vector database}
We used Facebook AI Similarity Search (FAISS)~\cite{douze2024faiss} to create and manage the vector database. The ten documents used, totaling approximately \num{2500} pages, resulting in \num{1525} chunks when divided into \num{3000}-character segments. We followed the FAISS documentation~\cite{douze2024faiss} and adopted flat indexing with a flat array data structure, as recommended for small databases ($|\mathcal{CV}|\leq \num{10000}$) where accuracy is a priority. 
%%%

\subsubsection{Similarity search} \label{sim}
We measured the similarity between query and document embeddings using \textit{Euclidean} distance (L2 norm), which represents the straight-line distance in the embedding space. We also use the default FAISS Heapsort method~\cite{schaffer1993analysis} for flat indexing~\cite{douze2024faiss}, efficiently retrieving the top $k=5$ most relevant chunks by selecting those with the smallest distances.
\subsubsection{LLM inference}
We selected four high-ranking open-source LLMs from the OpenLLM Leaderboard~\cite{open-llm-leaderboard-v1}, reflecting strong performance across diverse benchmarks, and used their pre-trained inferences for our experiments: Gemma2~\cite{team2024gemma} with \num{2} billion parameters and Llama3.1~\cite{touvron2023llama} with \num{8} billion parameters, Qwen2.5~\cite{qwen}, and Mistral~\cite{jiang2023mistral}, each with \num{7} billion parameters. Table~\ref{tab:llm} details the selected LLMs specifications. We prompt the LLM to analyze the user query and determine which tool ( FFmpeg, VVenC, or both) is required. During the first response-generation step, the model is instructed to synthesize a complete command using both the query and the retrieved documentation, providing all necessary parameters and a concise technical explanation of its functionality. In the self-reflection phase, the prompt directs the LLM to review its own output, checking command correctness, parameter completeness, and syntax formatting. To keep inference overhead minimal, we limit the maximum number of self-reflection iterations to one.

\subsection{Evaluation metrics}
We evaluate \texttt{ELLMPEG} accuracy, performance, and energy consumption using the following metrics.
\begin{enumerate}[topsep=0pt]
    \item\textit{Response accuracy} 
    follows established methodology~\cite{zheng2023judging} and evaluates the correctness of the generated commands using \emph{LLM-as-a-judge} based on GPT-4o~\cite{gpt40} and Gemini 2.0 Flash~\cite{gemini}. Each query-command pair can be \textit{a) correct:} if it fully aligns with the query and directly provides the expected command, or \textit{b) incorrect:} if it deviates from the query, lacks relevance, or misses key information. Accuracy is calculated as the percentage of correct responses across all queries.
    \item\textit{Response length} quantifies the total number of tokens in model responses, indicating the level of output detail.
    %%
    % \item\textit{Time to first token (TTFT)} 
    % measures initial response latency as the time from query submission to the receipt of the first response token.
    %%
%%%
    \item\textit{Tokens generated per second (TPS)} quantifies response generation speed as the total number of tokens generated divided by the inference time.
%%%
    \item\textit{Inference time} 
    measures the model's delay to generate a response from query submission to completion.
%%%%
    \item\textit{Inference energy consumption}
    estimates the energy consumption during inference (in \unit{\watt\hour}) using CodeCarbon~\cite{codecarbon_ref} based on the power draw of the system's CPU, GPU, and RAM during the inference period.
\end{enumerate}

We did not use metrics like BERTScore~\cite{zhang2019bertscore}, ROUGE~\cite{liu2008correlation}, and BLEU~\cite{papineni2002bleu}, which are not suitable for assessing FFmpeg and VVenC commands, as they focus on semantic similarity or token overlap, whereas command correctness depends on exact syntax. We also did not compare \texttt{ELLMPEG} to OpenAI-based LLMPEG~\cite{llmpeg}, as using GPT models would measure OpenAI's performance against itself, yielding no meaningful insights.

\begin{figure}[t]
    \centering      \includegraphics[width=0.95\linewidth]{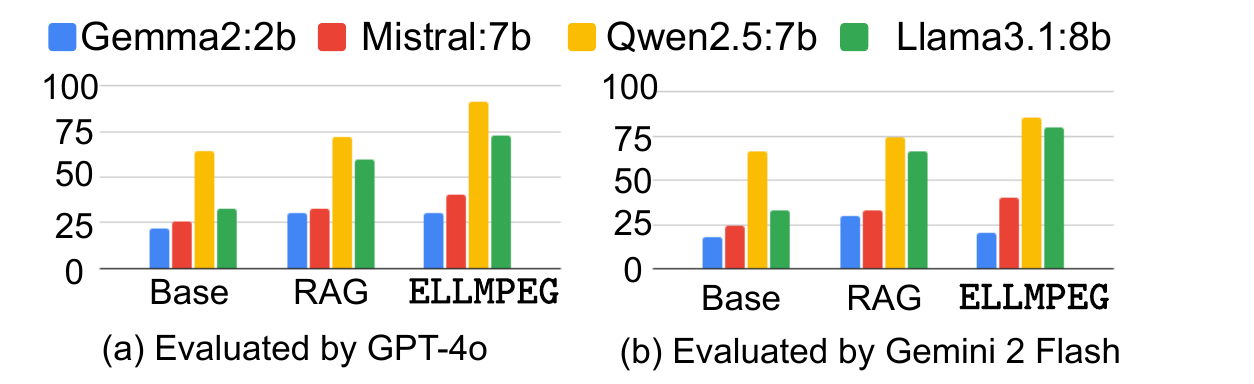}
    \caption{Accuracy on the FFmpeg command pool.}
    \label{fig:accffmpeg}
\end{figure}
\begin{figure}[t]
    \centering      \includegraphics[width=0.95\linewidth]{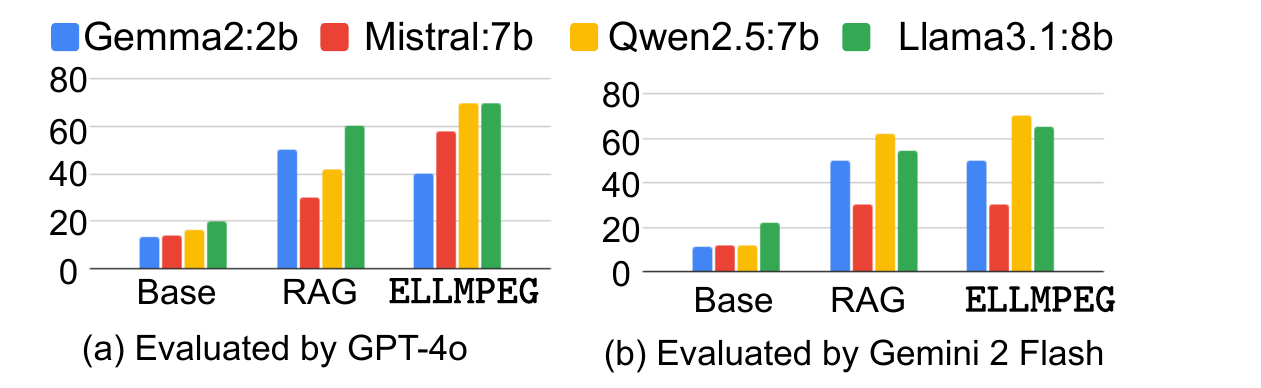}
    \caption{Accuracy on the VVenC command pool.}
    \label{fig:accvvc}
\end{figure}

\subsection{Baselines}
To evaluate the performance of \texttt{ELLMPEG}, we conduct an ablation study by progressively disabling key components to quantify their individual contributions. When running on edge hardware, using more than one self-reflection iteration ($I_{\max} > 1$) introduces significant computational overhead and substantially increases inference time; therefore, all edge experiments use $I_{\max} = 1$. We compare \texttt{ELLMPEG} against the following baseline configurations:
\begin{enumerate}[topsep=0pt]
    \item \textit{Base ablation} uses the underlying LLM without RAG retrieval or self-refinement, relying solely on its pre-trained knowledge to generate responses. Each query requires exactly one LLM inference.      
    \item \textit{RAG ablation} uses the LLM with tool-aware RAG and tool selection enabled, but without self-refinement, selecting the relevant tool(s), retrieving context from the corresponding vector store, and generating a response without iterative revision. This configuration requires two LLM calls, one for tool selection and one for response generation.
    \item \textit{Self-refinement depth ablation} evaluates the full \texttt{ELLMPEG} system with all components, response generation, and iterative self-refinement, using $I_{\max} \in\{1,2,3,4\}$. This experiment is executed only on the server due to the higher computational cost.
\end{enumerate}

%%%
\begin{figure}[!t]
    \centering
    \includegraphics[width=0.6\linewidth]{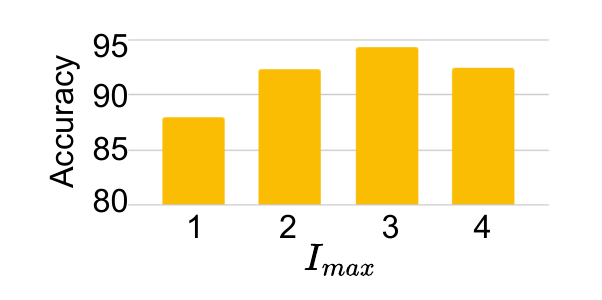}
    \caption{Accuracy of Qwen2.5 on FFmpeg queries under different self-reflection iteration depths}
    \label{fig:server}
\end{figure}
\section{Evaluation Results}
\label{sec:res}

This section evaluates \texttt{ELLMPEG} across all open-source models using the aforementioned metrics for the respective command pools. In addition to quantitative results, we demonstrate practical capabilities through visualizable examples. All results presented in this section represent the average values of each metric derived from all queries in the test set. 
%%%%%%%%
\subsection{Accuracy analysis} 
\begin{figure*}[!t]
    \centering
    \subfloat[Response length (tokens).]{%
        \includegraphics[width=0.5\textwidth]{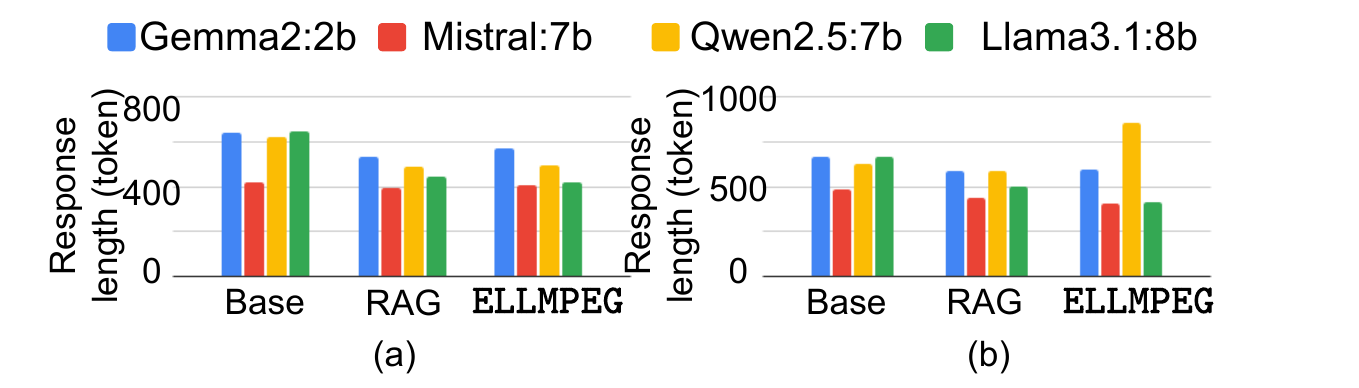}%
        \label{fig:length}
    }
    \subfloat[TPS ($\uparrow$ better).]{%
        \includegraphics[width=0.5\textwidth]{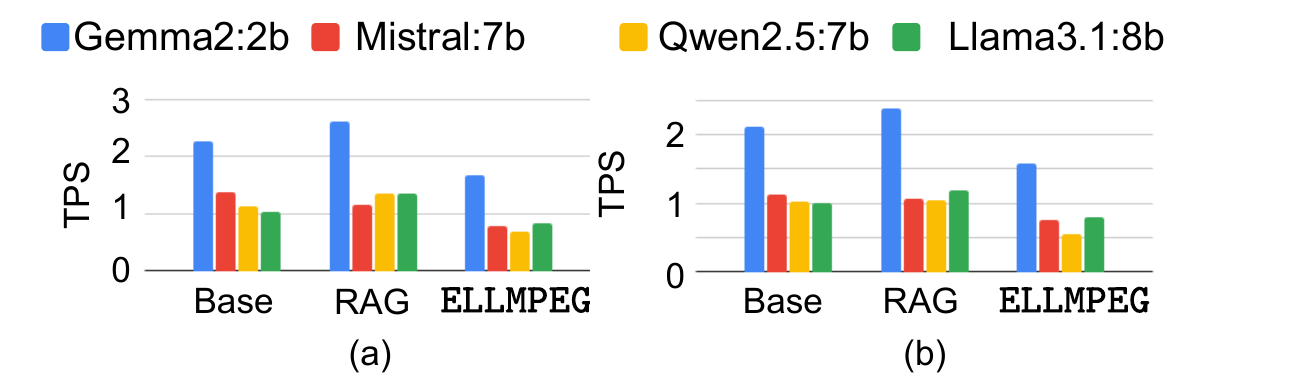}%
        \label{fig:tps}
    }    
    \caption{Average response length (tokens) and average TPS on FFmpeg and VVenC command pools.}
    \label{fig:tps+length}
\end{figure*}
\begin{figure*}[!t]
    \centering
    \subfloat[Inference time (\unit{\second}) ($\downarrow$ better).]{%
        \includegraphics[width=0.5\textwidth]{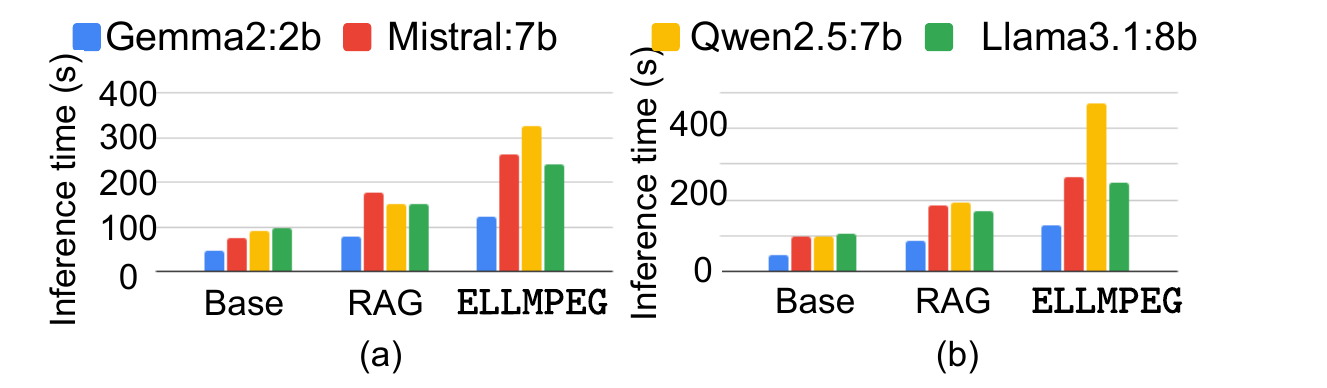}%
        \label{fig:time}
    }
    \subfloat[Inference energy (\unit{\watt\hour}) ($\downarrow$ better).]{%
        \includegraphics[width=0.5\textwidth]{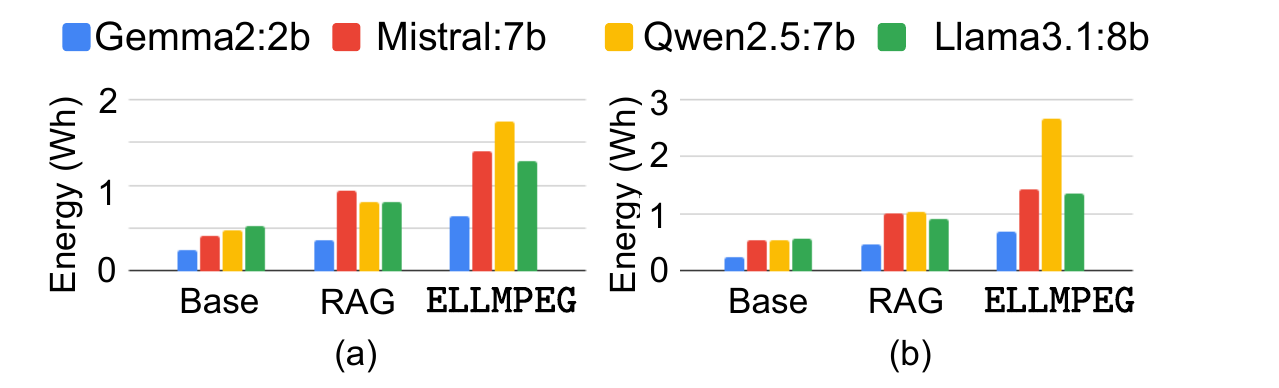}%
        \label{fig:energy}
    }    
    \caption{Average inference time and average inference energy consumption on FFmpeg and VVenC command pools.}
    \label{fig:time+energy}
\end{figure*}
Figs.~\ref{fig:accffmpeg} and~\ref{fig:accvvc} show \texttt{ELLMPEG}'s accuracy analysis across four open-source models for FFmpeg and VVenC queries, respectively. The analysis focuses on the following key aspects.
\subsubsection{Accuracy}\label{acc1}
Among all evaluated models, Qwen2.5 achieved the highest accuracy within the \texttt{ELLMPEG} architecture across both query pools. According to evaluations by GPT-4o and Gemini 2.0 Flash, Qwen2.5 reached \qty{88}{\percent} and \qty{85}{\percent} accuracy on the FFmpeg queries, and \qty{71}{\percent} and \qty{70}{\percent} on the VVenC queries.
Following Qwen2.5, Llama3.1-8B demonstrated solid performance, obtaining \qty{73}{\percent} and \qty{80}{\percent} accuracy on the FFmpeg query set, and \qty{70}{\percent} and \qty{65}{\percent} on VVenC.
In contrast, Gemma2, as the smallest model with only 2B parameters, does not benefit from the self-reflection module, as it lacks the capability to reliably evaluate its own responses. Mistral, despite having the same number of parameters as Qwen2.5 and being compatible with self-reflection, still produced relatively low accuracies, averaging \qty{40}{\percent} and \qty{44}{\percent}. 

Evaluation results indicate that performance differences across models integrated with \texttt{ELLMPEG} are also influenced by their ability to incorporate retrieved information. For example, Qwen2.5, with its large 128k-token context window, can more effectively integrate and process retrieved data, leading to higher accuracy. In contrast, models with smaller context windows (e.g., Gemma2 with 8k tokens and Mistral with 32k tokens) may truncate or overlook relevant portions of retrieved knowledge, resulting in reduced accuracy. When the retrieved content exceeds a model’s context capacity, the model struggles to form connections across different segments of the input, limiting its ability to generate correct and contextually grounded responses.
There is a strong consensus between GPT-4o and Gemini 2.0 Flash in judging FFmpeg command outputs, with only \qty{0.8}{\percent} deviation across all models and scenarios. On VVenC, the deviation is \qty{6.58}{\percent}, which is slightly higher but still indicates strong agreement between the two evaluators.

\subsection{Ablation analysis}
We conduct ablation studies on the \texttt{ELLMPEG} architecture to assess the roles of retrieval, self-reflection, and context handling. These analyses help isolate which factors drive the observed performance gains across FFmpeg and VVenC queries.

\subsubsection{Base ablation}
On FFmpeg queries, Qwen2.5 achieves the highest accuracy among the base models, with an average of \qty{65}{\percent}. The remaining models achieve only \qty{20}{\percent}–\qty{25}{\percent} on average. For VVenC queries, the average accuracy across all models drops to around \qty{15}{\percent}. This performance gap is expected; while the models have substantial built-in knowledge of FFmpeg due to its extensive public documentation and widespread usage, their familiarity with VVenC, a newer and more specialized encoder, is limited, as it is underrepresented in publicly available training data.

\subsubsection{RAG ablation}
Integrating the RAG module improves accuracy across all evaluated models, with particularly large gains on VVenC queries. This highlights that base models struggle with less commonly documented codecs and benefit significantly from retrieved domain-specific knowledge.
On FFmpeg queries, accuracy improvements with RAG are \qty{36}{\percent}, \qty{27}{\percent}, \qty{13}{\percent}, and \qty{81}{\percent} for Gemma2-2B, Mistral, Qwen2.5, and Llama3.1, respectively.
On VVenC queries, these improvements are substantially higher: \qty{280}{\percent}, \qty{110}{\percent}, \qty{160}{\percent}, and \qty{200}{\percent}, respectively.

Regarding tool selection, FFmpeg queries required only FFmpeg-related documentation, and all models correctly selected FFmpeg when prompted. For VVenC queries, however, model behavior diverged. Gemma and Mistral consistently selected VVenC, likely due to surface-level keyword matching rather than actual tool reasoning. Qwen2.5 and Llama3.1, in contrast, selected VVenC documentation in \qty{70}{\percent} of cases, and in the remaining \qty{30}{\percent}, they incorporated both FFmpeg and VVenC sources, suggesting a more deliberate interpretation of the query context.
\begin{table*}[!t]
    \centering
    \footnotesize
    \setlength{\tabcolsep}{1pt}
    \caption{Example outputs of \texttt{ELLMPEG} executable commands from the FFmpeg command pool.}
    \label{tab:execution}
    \begin{tabular}{|c|c|c|}
    \toprule
        Item & User query & Extracted executable command from \texttt{ELLMPEG}\\ \midrule
       \multirow{2}{*}{(a)}  & \multirow{2}{*}{How can I rotate a video by 90 degrees?} & \texttt{ffmpeg -i input.mp4 -vf }\\ 
       && \texttt{"transpose=cclock" output.mp4} \\
       \hline 
       \multirow{2}{*}{(b)}  & \multirow{2}{*}{How can I add letterboxing to a video?} & \texttt{ffmpeg -i input.mp4 -vf "scale=1280:720,}\\
       && \texttt{pad=1920:1080:(ow-iw)/2:(oh-ih)/2" output.mp4} \\
       \hline 
       \multirow{2}{*}{(c)}  & \multirow{2}{*}{How do I adjust the brightness and contrast of a video?} &  \texttt{ffmpeg -i input.mp4 -vf}  \\
       && \texttt{"eq=brightness=-10:contrast=+20" output.mp4} \\
       \hline 
       \multirow{2}{*}{(d)}  & \multirow{2}{*}{How do I add a logo to a video?} &  \texttt{ffmpeg -i input.mp4 -i logo.png }  \\
       && \texttt{-filter\_complex "overlay=W-w-10:H-h-10" output.mp4} \\
       \bottomrule
    \end{tabular}    
\end{table*}
\begin{figure*}[!t]
    \centering
\includegraphics[width=0.75\linewidth]{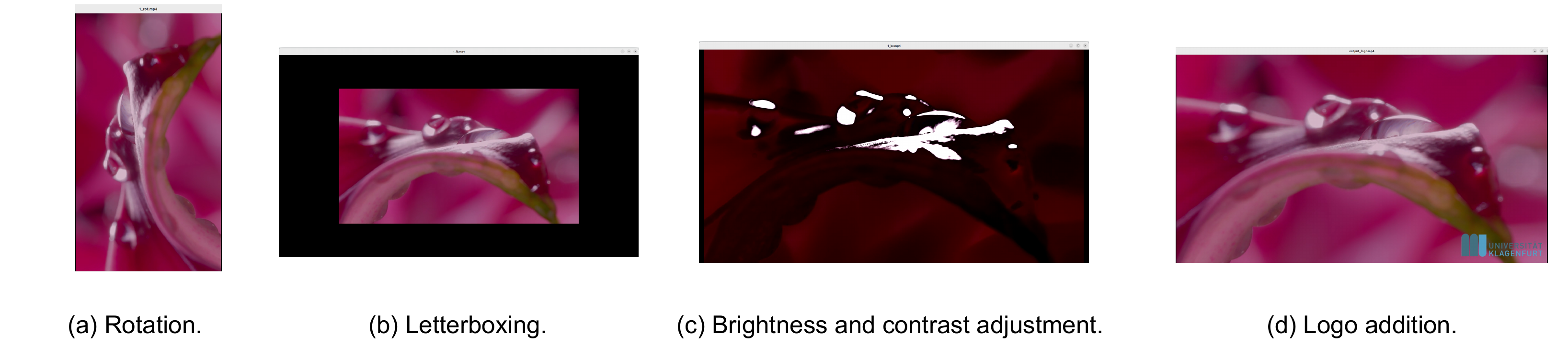}
    \caption{\texttt{ELLMPEG} execution results for the queries in Table~\ref{tab:execution}, played back using FFplay.}
    \label{fig:ex_res} 
\end{figure*}
\subsubsection{Self-reflection iteration depth ablation}
We additionally study the effect of increasing the number of self-reflection iterations in the \texttt{ELLMPEG} architecture. Because higher iteration counts introduce significant computational overhead, this experiment was executed on a server rather than the edge device. The evaluation is performed on Qwen2.5 using FFmpeg queries, with responses assessed by GPT-4.
The results in Fig.~\ref{fig:server} indicate that deeper self-reflection improves reasoning quality, but only up to a certain depth. In practice, three iterations strike a balanced trade-off between quality gains and computational cost. Beyond this depth, performance degrades as the model tends to over-correct valid outputs, hallucinating non-existent errors and introducing unnecessary modifications that ultimately compromise command validity.
\subsection{Performance analysis} 
Figs.~\ref{fig:tps+length} and \ref{fig:time+energy} show \texttt{ELLMPEG}'s performance across TPS, inference time, and energy consumption for the four open-source models when generating responses to FFmpeg and VVenC queries. We compare these results to the two ablations, without \texttt{ELLMPEG} enhancement. The following sections categorize our findings based on each metric.
\subsubsection{Response length}\label{resplen}
Fig.~\ref{fig:length} shows that, in most cases—except for Qwen2.5 on VVenC, the responses generated by \texttt{ELLMPEG} are on average \qty{12}{\percent} shorter. This reduction is particularly prominent for Llama3.1, which produces responses that are \qty{35}{\percent} shorter on FFmpeg queries and \qty{38}{\percent} shorter on VVenC queries. Compared to the RAG-only configuration, the response lengths in \texttt{ELLMPEG} remain more compact and generally similar. indicating that the architecture encourages concise, targeted outputs.

\subsubsection{Tokens generated per second (TPS)} 
Processing a longer input context and the additional computations introduced by self-reflection in \texttt{ELLMPEG} increase the overall computational load compared to the base and RAG models, resulting in slower response generation. As a result, TPS decreases for both FFmpeg and VVenC (Fig.~\ref{fig:tps}) with the average TPS dropping from \num{1.4} in the base models to roughly \num{1} in \texttt{ELLMPEG} for FFmpeg. For example, in Qwen2.5, the average TPS for the base model is \num{1.13}, which decreases to \num{0.69} with \texttt{ELLMPEG}, representing a \qty{40}{\percent} reduction. In comparison, Gemma2 experiences a smaller reduction of \qty{26}{\percent}, dropping from \num{2.27} in the base model to \num{1.67} with \texttt{ELLMPEG}, due to its smaller size.

\subsubsection{Inference time} 
Fig.~\ref{fig:time} shows the average inference time of base models, RAG, and \texttt{ELLMPEG} for both FFmpeg and VVenC test sets. Although the response length of \texttt{ELLMPEG} is on average shorter, the computational overhead of RAG and self-reflection results in higher inference times. While these values are relatively high, they are expected given the CPU-based evaluation setup. In \texttt{ELLMPEG}, since the inference time includes the time required for tool selection, RAG, first response generation, and self-reflection, it is expected to see higher inference time. The best performing model in terms of accuracy, Qwen2.5, comes with the highest inference time on both FFmpeg and VVenC datasets. The second-best performing shows lower inference time, with an acceptable level of accuracy.

\subsubsection{Inference energy} 
Fig.~\ref{fig:energy} compares the average energy consumption of base models and \texttt{ELLMPEG} for both FFmpeg and VVenC test sets. As shown for both test sets, \texttt{ELLMPEG} consumed more energy due to longer LLM inference. For FFmpeg queries, the average energy consumption of \texttt{ELLMPEG} increases by a factor of \num{1.7} for Gemma2, \num{2.5} for Mistral2, \num{2.6} for Qwen2.5, and \num{1.5} for Llama3.1, compared to their respective base models. On average, \qty{50}{\percent} of this increase comes from the RAG module, and the rest from the self-reflection overhead.

\subsection{Execution} 
We demonstrate the practical capabilities of \texttt{ELLMPEG} by selecting a set of visually representable commands from the command pool, such as video rotation and logo overlay. We use the input video sequence~\texttt{001} from the Inter4K dataset~\cite{stergiou2022adapool}. Table~\ref{tab:execution} presents the user queries along with the corresponding commands extracted from \texttt{ELLMPEG}'s responses, and Fig.~\ref{fig:ex_res} shows the resulting outputs.

\section{Conclusion}
\label{sec:conclusion}
This paper introduced \texttt{ELLMPEG}, an agentic LLM-based video-processing tool that combines a tool-aware RAG module, iterative self-reflection, and open-source edge-deployable LLMs to generate FFmpeg and VVenC commands from user queries. We created dedicated command pools for FFmpeg and VVenC and evaluated \texttt{ELLMPEG} using four LLMs across multiple metrics. Experimental results demonstrated that \texttt{ELLMPEG} significantly improves response accuracy by reducing incorrect answers across both test sets, achieving up to \qty{78}{\percent} accuracy in generating commands with zero recurring API costs. Our future directions include leveraging pre-trained domain-specific embeddings for improved accuracy, implementing query-level energy budgeting for dynamic inference optimization, and extending the work to GPU-based edge devices.

\begin{acks}
The financial support of the Austrian Federal Ministry for Digital and Economic Affairs, the National Foundation for Research, Technology and Development, and the Christian Doppler Research Association is gratefully acknowledged. Christian Doppler Laboratory ATHENA: \url{https://athena.itec.aau.at/}.
\end{acks}
\balance
\bibliographystyle{ieeetr}
\bibliography{mainbib}

@misc{llmpeg,
  author = {{Garrit Strenge}},
  title = {{LLMPEG}},
  year = {2024},
  howpublished = {\url{https://github.com/gstrenge/llmpeg}},
  note = {Accessed: 12 Nov. 2025}
}

@article{long2024videodrafter,
  title={{Videodrafter: Content-Consistent Multi-scene Video Generation With LLM}},
  author={Long, Fuchen and Qiu, Zhaofan and Yao, Ting and Mei, Tao},
  journal={arXiv preprint arXiv:2401.01256},
  year={2024}
}

@article{du2021glm,
  title={{GLM: General Language Model Pretraining With Autoregressive Blank Infilling}},
  author={Du, Zhengxiao and Qian, Yujie and Liu, Xiao and Ding, Ming and Qiu, Jiezhong and Yang, Zhilin and Tang, Jie},
  journal={arXiv preprint arXiv:2103.10360},
  year={2021}
}

@article{fu2024video,
  title={{Video-MME: The First-Ever Comprehensive Evaluation Benchmark of Multi-modal LLMs in Video Analysis}},
  author={Fu, Chaoyou and Dai, Yuhan and Luo, Yondong and Li, Lei and Ren, Shuhuai and Zhang, Renrui and Wang, Zihan and Zhou, Chenyu and Shen, Yunhang and Zhang, Mengdan and others},
  journal={arXiv preprint arXiv:2405.21075},
  year={2024}
}

@inproceedings{wang2024lave,
  title={{LAVE: LLM-Powered Agent Assistance and Language Augmentation for Video Editing}},
  author={Wang, Bryan and Li, Yuliang and Lv, Zhaoyang and Xia, Haijun and Xu, Yan and Sodhi, Raj},
  booktitle={{Proceedings of the 29th International Conference on Intelligent User Interfaces}},  
  year={2024}
}

@article{huang2024free,
  title={{Free-Bloom: Zero-Shot Text-to-Video Generator with LLM Director and LDM Animator}},
  author={Huang, Hanzhuo and Feng, Yufan and Shi, Cheng and Xu, Lan and Yu, Jingyi and Yang, Sibei},
  journal={{Advances in Neural Information Processing Systems}},  
  year={2024}
}

@misc{openai,
  author = {{Open AI}},
  title = {{The Most Powerful Platform for Building AI Products}},
  year = {2024},
  howpublished = {\url{https://openai.com/api/}},
  note = {Accessed: 12 Nov. 2025}
}

@misc{ffmpeg,
  author = {{FFmpeg}},
  title = {{A Complete, Cross-Platform Solution to Record, Convert and Stream Audio and Video.}},
  year = {2024},
  howpublished = {\url{https://www.ffmpeg.org/}},
  note = {Accessed: 12 Nov. 2025}
}

@misc{codecarbon_ref,
  author = {{CodeCarbon Development Team}},
  title = {{Track and Reduce CO2 Emissions From Your Computing}},
  year = {2024},
  howpublished = {\url{https://codecarbon.io/}},
  note = {Accessed: 12 Nov. 2025}
}

@article{jiang2023mistral,
  title={{Mistral 7B}},
  author={Jiang, Albert Q and Sablayrolles, Alexandre and Mensch, Arthur and Bamford, Chris and Chaplot, Devendra Singh and Casas, Diego de las and Bressand, Florian and Lengyel, Gianna and Lample, Guillaume and Saulnier, Lucile and others},
  journal={arXiv preprint arXiv:2310.06825},
  year={2023}
}

@article{team2024gemma,
  title={{Gemma 2: Improving Open Language Models at a Practical Size}},
  author={Team, Gemma and Riviere, Morgane and Pathak, Shreya and Sessa, Pier Giuseppe and Hardin, Cassidy and Bhupatiraju, Surya and Hussenot, L{\'e}onard and Mesnard, Thomas and Shahriari, Bobak and Ram{\'e}, Alexandre and others},
  journal={arXiv preprint arXiv:2408.00118},
  year={2024}
}

@misc{qwen,
  author = {{Qwen}},
  title = {{Qwen2.5}},
  year = {2024},
  howpublished = {\url{https://huggingface.co/Qwen/Qwen2.5-7B}},
  note = {Accessed: 12 Nov. 2025}
}

@article{lin2024streamingbench,
  title={{StreamingBench: Assessing the Gap for MLLMs to Achieve Streaming Video Understanding}},
  author={Lin, Junming and Fang, Zheng and Chen, Chi and Wan, Zihao and Luo, Fuwen and Li, Peng and Liu, Yang and Sun, Maosong},
  journal={arXiv preprint arXiv:2411.03628},
  year={2024}
}

@article{douze2024faiss,
  title={{The FAISS Library}},
  author={Douze, Matthijs and Guzhva, Alexandr and Deng, Chengqi and Johnson, Jeff and Szilvasy, Gergely and Mazaré, Pierre-Emmanuel and Lomeli, Maria and Hosseini, Lucas and Jégou, Hervé},
  journal={arXiv preprint arXiv:2401.08281},
  year={2024},
  }

@misc{open-llm-leaderboard-v1,
  author = {Beeching, Edward and Fourrier, Clémentine and Habib, Nathan and Han, Sheon and Lambert, Nathan and Rajani, Nazneen and Sanseviero, Omar and Tunstall, Lewis and Wolf, Thomas},
  title = {{Open LLM Leaderboard}},
  year = {2023},
  publisher = {{Hugging Face}},
  howpublished = "\url{https://huggingface.co/spaces/open-llm-leaderboard-old/open_llm_leaderboard}",
  note = {Accessed: 12 Nov. 2025}
}

@article{zheng2023judging,
  title={{Judging LLM-as-a-Judge with MT-Bench and Chatbot Arena}},
  author={Zheng, Lianmin and Chiang, Wei-Lin and Sheng, Ying and Zhuang, Siyuan and Wu, Zhanghao and Zhuang, Yonghao and Lin, Zi and Li, Zhuohan and Li, Dacheng and Xing, Eric and others},
  journal={{Advances in Neural Information Processing Systems}},  
  year={2023}
}

@misc{gpt40,
  author = {{OpenAI}},
  title = {{GPT-4o}},
  year = {2023},
  howpublished = "\url{https://openai.com/index/hello-gpt-4o/}",
  note = {Accessed: 12 Nov. 2025}
}

@article{zhang2019bertscore,
  title={{Bertscore: Evaluating Text Generation with Bert}},
  author={Zhang, Tianyi and Kishore, Varsha and Wu, Felix and Weinberger, Kilian Q and Artzi, Yoav},
  journal={arXiv preprint arXiv:1904.09675},
  year={2019}
}

@inproceedings{liu2008correlation,
  title={{Correlation Between Rouge and Human Evaluation of Extractive Meeting Summaries}},
  author={Liu, Feifan and Liu, Yang},
  booktitle={{Proceedings of ACL-08: HLT}},  
  year={2008}
}

@inproceedings{papineni2002bleu,
  title={{BLEU: A Method For Automatic Evaluation of Machine Translation}},
  author={Papineni, Kishore and Roukos, Salim and Ward, Todd and Zhu, Wei-Jing},
  booktitle={{Proceedings of the 40th Annual Meeting of the Association for Computational Linguistics}},  
  year={2002}
}

@misc{bge_embedding,
  title={{C-Pack: Packaged Resources To Advance General Chinese Embedding}}, 
  author={Xiao, Shitao and Liu, Zheng and Zhang, Peitian and Muennighoff, Niklas},
  journal={arXiv preprint arXiv:2309.07597},
  year={2023},
  }

@article{zhang2023simple,
  title={{A Simple LLM Framework for Long-range Video Question-Answering}},
  author={Zhang, Ce and Lu, Taixi and Islam, Md Mohaiminul and Wang, Ziyang and Yu, Shoubin and Bansal, Mohit and Bertasius, Gedas},
  journal={arXiv preprint arXiv:2312.17235},
  year={2023}
}

@inproceedings{pan2023retrieving,
  title={{Retrieving-to-Answer: Zero-Shot Video Question Answering with Frozen Large Language Models}},
  author={Pan, Junting and Lin, Ziyi and Ge, Yuying and Zhu, Xiatian and Zhang, Renrui and Wang, Yi and Qiao, Yu and Li, Hongsheng},
  booktitle={{Proceedings of the IEEE/CVF International Conference on Computer Vision}},  
  year={2023}
}

@inproceedings{arefeen2024vita,
  title={{ViTA: An Efficient Video-to-Text Algorithm using VLM for RAG-based Video Analysis System}},
  author={Arefeen, Md Adnan and Debnath, Biplob and Uddin, Md Yusuf Sarwar and Chakradhar, Srimat},
  booktitle={{Proceedings of the IEEE/CVF Conference on Computer Vision and Pattern Recognition}},  
  year={2024}
}

@inproceedings{arefeen2024irag,
  title={{iRAG: Advancing RAG for Videos with an Incremental Approach}},
  author={Arefeen, Md Adnan and Debnath, Biplob and Uddin, Md Yusuf Sarwar and Chakradhar, Srimat},
  booktitle={{Proceedings of the 33rd ACM International Conference on Information and Knowledge Management}},  
  year={2024}
}

@article{luo2024video,
  title={{Video-RAG: Visually-aligned Retrieval-Augmented Long Video Comprehension}},
  author={Luo, Yongdong and Zheng, Xiawu and Yang, Xiao and Li, Guilin and Lin, Haojia and Huang, Jinfa and Ji, Jiayi and Chao, Fei and Luo, Jiebo and Ji, Rongrong},
  journal={arXiv preprint arXiv:2411.13093},
  year={2024}
}

@article{touvron2023llama,
  title={{Llama: Open and Efficient Foundation Language Models}},
  author={Touvron, Hugo and Lavril, Thibaut and Izacard, Gautier and Martinet, Xavier and Lachaux, Marie-Anne and Lacroix, Timoth{\'e}e and Rozi{\`e}re, Baptiste and Goyal, Naman and Hambro, Eric and Azhar, Faisal and others},
  journal={arXiv preprint arXiv:2302.13971},
  year={2023}
}

@misc{TextSplitter,
  author = {LangChain},
  title = {{Langchain Text Splitters}},
  year = {2024},  
  howpublished = "\url{https://python.langchain.com/api_reference/text_splitters/index.html}", 
  note = {Accessed: 12 Nov. 2025}
}

@misc{tokentochar,
  author = {{OpenAI}},
  title = {{What are Tokens and How to Count Them?}},
  year = {2024},  
  howpublished = "\url{https://help.openai.com/en/articles/4936856-what-are-tokens-and-how-to-count-them}", 
  note = {Accessed: 12 Nov. 2025}
}

@misc{ffmpegDoc,
  author = {{FFmpeg}},
  title = {{FFmpeg Documentation}},
  year = {2024},  
  howpublished = "\url{https://ffmpeg.org/documentation.html}", 
  note = {Accessed: 12 Nov. 2025}
}

@misc{vvcDoc,
  author = {{Fraunhofer HHI}},
  title = {{VVenC}},
  year = {2024},  
  howpublished = "\url{https://github.com/fraunhoferhhi/vvenc/wiki}", 
  note = {Accessed: 12 Nov. 2025}
}

@article{schaffer1993analysis,
  title={{The Analysis of Heapsort}},
  author={Schaffer, Russel and Sedgewick, Robert},
  journal={{Journal of Algorithms}},
  year={1993},
  publisher={Elsevier}
}

@article{farahani2024towards,
  title={{Towards AI-Assisted Sustainable Adaptive Video Streaming Systems: Tutorial and Survey}},
  author={Farahani, Reza and Azimi, Zoha and Timmerer, Christian and Prodan, Radu},
  journal={arXiv preprint arXiv:2406.02302},
  year={2024}
}

@inproceedings{farahani2023sarena,
  title={{SARENA: SFC-enabled Architecture for Adaptive Video Streaming Applications}},
  author={Farahani, Reza and Bentaleb, Abdelhak and Timmerer, Christian and Shojafar, Mohammad and Prodan, Radu and Hellwagner, Hermann},
  booktitle={ICC 2023-IEEE International Conference on Communications},
  pages={864--870},
  year={2023},
  organization={IEEE}
}

@inproceedings{farahani2022richter,
  title={{RICHTER: Hybrid P2P-CDN Architecture for Low Latency Live Video Streaming}},
  author={Farahani, Reza and Amirpour, Hadi and Tashtarian, Farzad and Bentaleb, Abdelhak and Timmerer, Christian and Hellwagner, Hermann and Zimmermann, Roger},
  booktitle={Proceedings of the 1st Mile-High Video Conference},
  year={2022}
}

@article{lin2023video,
  title={{Video-LLaVA: Learning United Visual Representation by Alignment Before Projection}},
  author={Lin, Bin and Ye, Yang and Zhu, Bin and Cui, Jiaxi and Ning, Munan and Jin, Peng and Yuan, Li},
  journal={arXiv preprint arXiv:2311.10122},
  year={2023}
}

@article{wang2024qwen2,
  title={{Qwen2-VL: Enhancing Vision-Language Model's Perception of the World at any Resolution}},
  author={Wang, Peng and Bai, Shuai and Tan, Sinan and Wang, Shijie and Fan, Zhihao and Bai, Jinze and Chen, Keqin and Liu, Xuejing and Wang, Jialin and Ge, Wenbin and others},
  journal={arXiv preprint arXiv:2409.12191},
  year={2024}
}

@inproceedings{huang2024vtimellm,
  title={{VTimeLLM: Empower LLM to Grasp Video Moments}},
  author={Huang, Bin and Wang, Xin and Chen, Hong and Song, Zihan and Zhu, Wenwu},
  booktitle={{Proceedings of the IEEE/CVF Conference on Computer Vision and Pattern Recognition}},
  year={2024}
}

@misc{vicuna2023,
    title = {{Vicuna: An Open-Source Chatbot Impressing GPT-4 with 90\%* ChatGPT Quality}},
    howpublished = "\url{https://lmsys.org/blog/2023-03-30-vicuna/}",
    author = {Chiang, Wei-Lin and Li, Zhuohan and Lin, Zi and Sheng, Ying and Wu, Zhanghao and Zhang, Hao and Zheng, Lianmin and Zhuang, Siyuan and Zhuang, Yonghao and Gonzalez, Joseph E. and Stoica, Ion and Xing, Eric P.},
    year = {2023}
}

@article{lin2023videodirectorgpt,
  title={{Videodirectorgpt: Consistent Multi-scene Video Generation via LLM-guided Planning}},
  author={Lin, Han and Zala, Abhay and Cho, Jaemin and Bansal, Mohit},
  journal={arXiv preprint arXiv:2309.15091},
  year={2023}
}

@article{liu2023visual,
  title={{Visual Instruction Tuning}},
  author={Liu, Haotian and Li, Chunyuan and Wu, Qingyang and Lee, Yong Jae},
  journal={{Advances in Neural Information Processing Systems}},
  year={2023}
}

@misc{sandvine,
  author = {{AppLogic Networks}},
  title = {{2024 Global Internet Phenomena Report}},
  year = {2024},
  howpublished = {\url{https://www.sandvine.com/}},
  note = {Accessed: 12 Nov. 2025}
}

@misc{gitffmpeg,
  author = {{FFmpeg}},
  title = {{FFmpeg}},
  year = {2024},
  howpublished = {\url{https://github.com/FFmpeg/FFmpeg}},
  note = {Accessed: 12 Nov. 2025}
}

@misc{wikiffmpeg,
  author = {{AMIAopensource}},
  title = {{FFmprovisr}},
  year = {2024},
  howpublished = {\url{https://amiaopensource.github.io/ffmprovisr/}},
  note = {Accessed: 12 Nov. 2025}
}

@misc{wikivvc,
  author = {{Rosato, Gianni and others}},
  title = {{VVenC}},
  year = {2024},
  howpublished = {\url{https://wiki.x266.mov/docs/encoders/VVenC}},
  note = {Accessed: 12 Nov. 2025}
}

@misc{gitvvc1,
  author = {{Fraunhofer HHI}},
  title = {{FFmpeg Integration}},
  year = {2024},
  howpublished = {\url{https://github.com/fraunhoferhhi/vvenc/wiki/FFmpeg-Integration}},
  note = {Accessed: 12 Nov. 2025}
}

@misc{gitvvc2,
  author = {{Fraunhofer HHI}},
  title = {{VVenC}},
  year = {2024},
  howpublished = {\url{https://github.com/fraunhoferhhi/vvenc}},
  note = {Accessed: 12 Nov. 2025}
}

@inproceedings{aggarwal2025programming,
  title={{Programming with Pixels: Towards Generalist Software Engineering Agents}},
  author={Aggarwal, Pranjal and Welleck, Sean},
year={2025},
  booktitle={{ICLR 2025 Third Workshop on Deep Learning for Code}}
}

@article{cao2024reframe,
  title={{Reframe Anything: LLM Agent for Open World Video Reframing}},
  author={Cao, Jiawang and Wu, Yongliang and Chi, Weiheng and Zhu, Wenbo and Su, Ziyue and Wu, Jay},
  journal={arXiv preprint arXiv:2403.06070},
  year={2024}
}

@misc{gemini,
  author = {{Google}},
  title = {{Gemini 2.0 Flash}},
  year = {2025},
  howpublished = {\url{https://ai.google.dev/gemini-api/docs/models#gemini-2.0-flash}},
  note = {Accessed: 12 Nov. 2025}
}

@inproceedings{ahmed2024codeqa,
  title={{CodeQA: Advanced Programming Question-Answering Using LLM Agent and RAG}},
  author={Ahmed, Mohamed and Dorrah, Mostafa and Ashraf, Ahmed and Adel, Yousef and Elatrozy, Abdelrahman and Mohamed, Bahaa Eldin and Gomaa, Wael},
  booktitle={{6th Novel Intelligent and Leading Emerging Sciences Conference (NILES)}},
  year={2024},
  organization={{IEEE}}
}

@inproceedings{alario2024tailoring,
  title={{Tailoring Your Code Companion: Leveraging LLMs and RAG to Develop a Chatbot to Support Students in a Programming Course}},
  author={Alario-Hoyos, Carlos and Kemcha, Rebiha and Kloos, Carlos Delgado and Callejo, Patricia and Est{\'e}vez-Ayres, Iria and Sant{\'\i}n-Crist{\'o}bal, David and Cruz-Argudo, Francisco and L{\'o}pez-S{\'a}nchez, Jos{\'e} Luis},
  booktitle={{IEEE International Conference on Teaching, Assessment and Learning for Engineering (TALE)}},
  year={2024},
  organization={{IEEE}}
}

@inproceedings{nam2024using,
  title={{Using an LLM to Help with Code Understanding}},
  author={Nam, Daye and Macvean, Andrew and Hellendoorn, Vincent and Vasilescu, Bogdan and Myers, Brad},
  booktitle={{Proceedings of the IEEE/ACM 46th International Conference on Software Engineering}},
  year={2024}
}

@article{stergiou2022adapool,
  title={Adapool: Exponential Adaptive Pooling for Information-retaining Downsampling},
  author={Stergiou, Alexandros and Poppe, Ronald},
  journal={{IEEE Transactions on Image Processing}},
  year={2022},
  publisher={{IEEE}}
}

@article{yanava,
  title={{AVA: Towards Agentic Video Analytics with Vision Language Models}},
  author={Yan, Yuxuan and Jiang, Shiqi and Cao, Ting and Yang, Yifan and Yang, Qianqian and Shu, Yuanchao and Yang, Yuqing and Qiu, Lili},
  year = {2025},
}

@article{tran2025towards,
  title={{Towards Agentic AI for Multimodal-Guided Video Object Segmentation}},
  author={Tran, Tuyen and Le, Thao Minh and Tran, Truyen},
  journal={arXiv preprint arXiv:2508.10572},
  year={2025}
}

@inproceedings{liu2024grounding,
  title={{Grounding Dino: Marrying Dino with Grounded Pre-training for Open-set Object Detection}},
  author={Liu, Shilong and Zeng, Zhaoyang and Ren, Tianhe and Li, Feng and Zhang, Hao and Yang, Jie and Jiang, Qing and Li, Chunyuan and Yang, Jianwei and Su, Hang and others},
  booktitle={{European Conference on Computer Vision}},
  year={2024},
  organization={{Springer}}
}

@article{ravi2024sam,
  title={{Sam 2: Segment Anything in Images and Videos}},
  author={Ravi, Nikhila and Gabeur, Valentin and Hu, Yuan-Ting and Hu, Ronghang and Ryali, Chaitanya and Ma, Tengyu and Khedr, Haitham and R{\"a}dle, Roman and Rolland, Chloe and Gustafson, Laura and others},
  journal={arXiv preprint arXiv:2408.00714},
  year={2024}
}

@article{chen2022beats,
  title={{Beats: Audio Pre-training with Acoustic Tokenizers}},
  author={Chen, Sanyuan and Wu, Yu and Wang, Chengyi and Liu, Shujie and Tompkins, Daniel and Chen, Zhuo and Wei, Furu},
  journal={arXiv preprint arXiv:2212.09058},
  year={2022}
}

@misc{llama3.3,
  author = {{Meta-llama}},
  title = {{Meta Llama 3.3 Multilingual Large Language Model}},
  year = {2024},
  howpublished = {\url{https://huggingface.co/meta-llama/Llama-3.3-70B-Instruct}},
  note = {Accessed: 12 Nov. 2025}
}

@article{zhang2025deep,
  title={{Deep Video Discovery: Agentic Search with Tool Use for Long-form Video Understanding}},
  author={Zhang, Xiaoyi and Jia, Zhaoyang and Guo, Zongyu and Li, Jiahao and Li, Bin and Li, Houqiang and Lu, Yan},
  journal={arXiv preprint arXiv:2505.18079},
  year={2025}
}

@article{ding2025prompt,
  title={{Prompt-Driven Agentic Video Editing System: Autonomous Comprehension of Long-Form, Story-Driven Media}},
  author={Ding, Zihan and Wang, Xinyi and Chen, Junlong and Kristensson, Per Ola and Shen, Junxiao},
  journal={arXiv preprint arXiv:2509.16811},
  year={2025}
}

@article{shinn2023reflexion,
  title={{Reflexion: Language Agents with Verbal Reinforcement Learning}},
  author={Shinn, Noah and Cassano, Federico and Gopinath, Ashwin and Narasimhan, Karthik and Yao, Shunyu},
  journal={{Advances in Neural Information Processing Systems}},
  year={2023}
}

@article{gou2023critic,
  title={{Critic: Large language models can self-correct with tool-interactive critiquing}},
  author={Gou, Zhibin and Shao, Zhihong and Gong, Yeyun and Shen, Yelong and Yang, Yujiu and Duan, Nan and Chen, Weizhu},
  journal={arXiv preprint arXiv:2305.11738},
  year={2023}
}

@article{madaan2023self,
  title={{Self-refine: Iterative Refinement with Self-feedback}},
  author={Madaan, Aman and Tandon, Niket and Gupta, Prakhar and Hallinan, Skyler and Gao, Luyu and Wiegreffe, Sarah and Alon, Uri and Dziri, Nouha and Prabhumoye, Shrimai and Yang, Yiming and others},
  journal={{Advances in Neural Information Processing Systems}},
  year={2023}
}

@article{belcak2025small,
  title={{Small Language Models are the Future of Agentic AI}},
  author={Belcak, Peter and Heinrich, Greg and Diao, Shizhe and Fu, Yonggan and Dong, Xin and Muralidharan, Saurav and Lin, Yingyan Celine and Molchanov, Pavlo},
  journal={arXiv preprint arXiv:2506.02153},
  year={2025}
}

@article{xi2025rise,
  title={{The Rise and Potential of Large Language Model Based Agents: A Survey}},
  author={Xi, Zhiheng and Chen, Wenxiang and Guo, Xin and He, Wei and Ding, Yiwen and Hong, Boyang and Zhang, Ming and Wang, Junzhe and Jin, Senjie and Zhou, Enyu and others},
  journal={{Science China Information Sciences}},
  year={2025},
  publisher={{Springer}}
}

@article{sullivan2012overview,
  title={{Overview of the High Efficiency Video Coding (HEVC) Standard}},
  author={Sullivan, G J and others},
  journal={{Tran. on Circuits and Systems for Video Technology}}, 
  year={2012},
  publisher={{IEEE}}
}

@article{wiegand2003overview,
  title={{Overview of the H. 264/AVC Video Coding Standard}},
  author={Wiegand, T and others},
  journal={{Tran. on Circuits and Systems for Video Technology}},
  year={2003},
  publisher={{IEEE}}
}

@inproceedings{azimi2025towards,
  title={{Towards an Energy-Efficient Video Processing Tool with LLMs}},
  author={Azimi, Zoha and Farahani, Reza and Timmerer, Christian and Prodan, Radu},
  booktitle={{Proc. of the 4th Mile-High Video Conf.}},
  year={2025}
}

@inproceedings{menon2023transcoding,
  title={{Transcoding quality prediction for adaptive video streaming}},
  author={Menon, Vignesh V and Farahani, Reza and Rajendran, Prajit T and Ghanbari, Mohammed and Hellwagner, Hermann and Timmerer, Christian},
  booktitle={Proceedings of the 2nd Mile-High Video Conference},
  year={2023}
}

@inproceedings{menon2023energy,
  title={Energy-Efficient Multi-Codec Bitrate-Ladder Estimation for Adaptive Video Streaming},
  author={Menon, Vignesh V and Farahani, Reza and Rajendran, Prajit T and Afzal, Samira and Schoeffmann, Klaus and Timmerer, Christian},
  booktitle={2023 IEEE International Conference on Visual Communications and Image Processing (VCIP)},
  year={2023},
  organization={IEEE}
}
\vspace{12pt}
\end{document}